\documentclass[10pt,journal,compsoc,twocolumn]{IEEEtran}

\usepackage{booktabs}
\usepackage{multirow}
\usepackage{ragged2e}
\usepackage{hyperref}
\usepackage{dsfont,amsfonts,amssymb,amsmath,color}
\usepackage{latexsym}
\usepackage{amsmath}
\usepackage{amsfonts}
\usepackage{algorithm}
\usepackage[noend]{algpseudocode}
\usepackage{graphicx}
\usepackage{subfigure}
\usepackage{color}
\usepackage{array}
\usepackage{algorithm}  
\usepackage{algpseudocode}
\usepackage{amsmath}  
\usepackage{makecell}
\usepackage{cases}
\usepackage{color}
\usepackage{amsthm}
\usepackage{bm}

\usepackage{soul}
\usepackage{tabularx}
\usepackage{makecell}
\usepackage{mwe}
\usepackage{fixltx2e}

\newtheorem{problem}{Problem}[section]
\newtheorem{definition}{Definition}[section]

\makeatletter
\def\hlinewd#1{%
  \noalign{\ifnum0=`}\fi\hrule \@height #1 \futurelet
   \reserved@a\@xhline}
\makeatother

\begin{document}

\title{Secure Your Ride: Real-time Matching Success Rate Prediction for Passenger-Driver Pairs}

\author{c\\
	% <-this % stops a space
	\IEEEcompsocitemizethanks{
		\IEEEcompsocthanksitem Y. Wang is with the School of Computer Science and Engineering, Beihang University, Beijing, China. \protect\\ 
		E-mail: wangyd@act.buaa.edu.cn
		\IEEEcompsocthanksitem H. Yin and T. Chen are with the School of Information Technology and Electrical Engineering, The University of Queensland, Brisbane, Australia\protect\\
		E-mail: \{h.yin1, tong.chen\}@uq.edu.au
		\IEEEcompsocthanksitem L. Wu and C. Liu are with the Department of Trading Engine, Didi Chuxing, Beijing, China\protect\\
		E-mail: \{wulian, liuchunyang\}@didiglobal.com
}
\thanks{Manuscript received March XX, 2021. Corresponding author: Hongzhi Yin.}
}

%\markboth{}%
%{Shell \MakeLowercase{\textit{et al.}}: Bare Demo of IEEEtran.cls for Computer Society Journals}

%==================================================================================
\IEEEtitleabstractindextext{%
\begin{abstract}
\justifying
In recent years, online ride-hailing platforms, such as Uber and Didi, have become an indispensable part of urban transportation and make our lives more convenient. After a passenger is matched up with a driver by the platform, both the passenger and the driver have the freedom to simply accept or cancel a ride with one click. 
Hence, accurately predicting whether a passenger-driver pair is a good match, i.e., its matching success rate (MSR), turns out to be crucial for ride-hailing platforms to devise instant strategies such as order assignment. 
However, since the users of ride-hailing platforms consist of two parties, decision-making needs to simultaneously account for the dynamics from both the driver and the passenger sides. This makes it more challenging than traditional online advertising tasks that predict a user's response towards an object, e.g., click-through rate prediction for advertisements. Moreover, the amount of available data is severely imbalanced across different cities, creating difficulties for training an accurate model for smaller cities with scarce data. Though a sophisticated neural network architecture can help improve the prediction accuracy under data scarcity, the overly complex design will impede the model's capacity of delivering timely predictions in a production environment. In the paper, to accurately predict the MSR of passenger-driver, we propose the \underline{\textbf{M}}ulti-\underline{\textbf{V}}iew model (\textbf{MV}) which comprehensively learns the interactions among the dynamic features of the passenger, driver, trip order, as well as the context. Regarding the data imbalance problem, we further design the \underline{\textbf{K}}nowledge \underline{\textbf{D}}istillation framework (\textbf{KD}) to supplement the model's predictive power for smaller cities using the knowledge from cities with denser data, and also generate a simple model to support efficient deployment. Finally, we conduct extensive experiments on real-world datasets from several different cities, which demonstrates the superiority of our solution. 
\end{abstract}

}

%==========================================================================================

\maketitle

\IEEEdisplaynontitleabstractindextext

\IEEEpeerreviewmaketitle

\section{Introduction}\label{sec:introduction} 
With the prominent development of technologies like GPS and mobile Internet, various ride-hailing applications have been prosperous in providing drivers and passengers with more convenience, such as Didi, Lyft and Uber. In every ride-hailing platform, when travel orders emerge, services are offered by matching passengers with drivers in a timely manner. Hence, for those platforms, predicting the matching success rate of a candidate passenger-driver pair plays an important role when making efficient order assignments.
\begin{figure}[tb!]
	\centering
	\setlength{\abovecaptionskip}{0.10cm}
	\subfigure[]{
		\label{fig:intro:sub1}
		\includegraphics[width=0.23\textwidth]{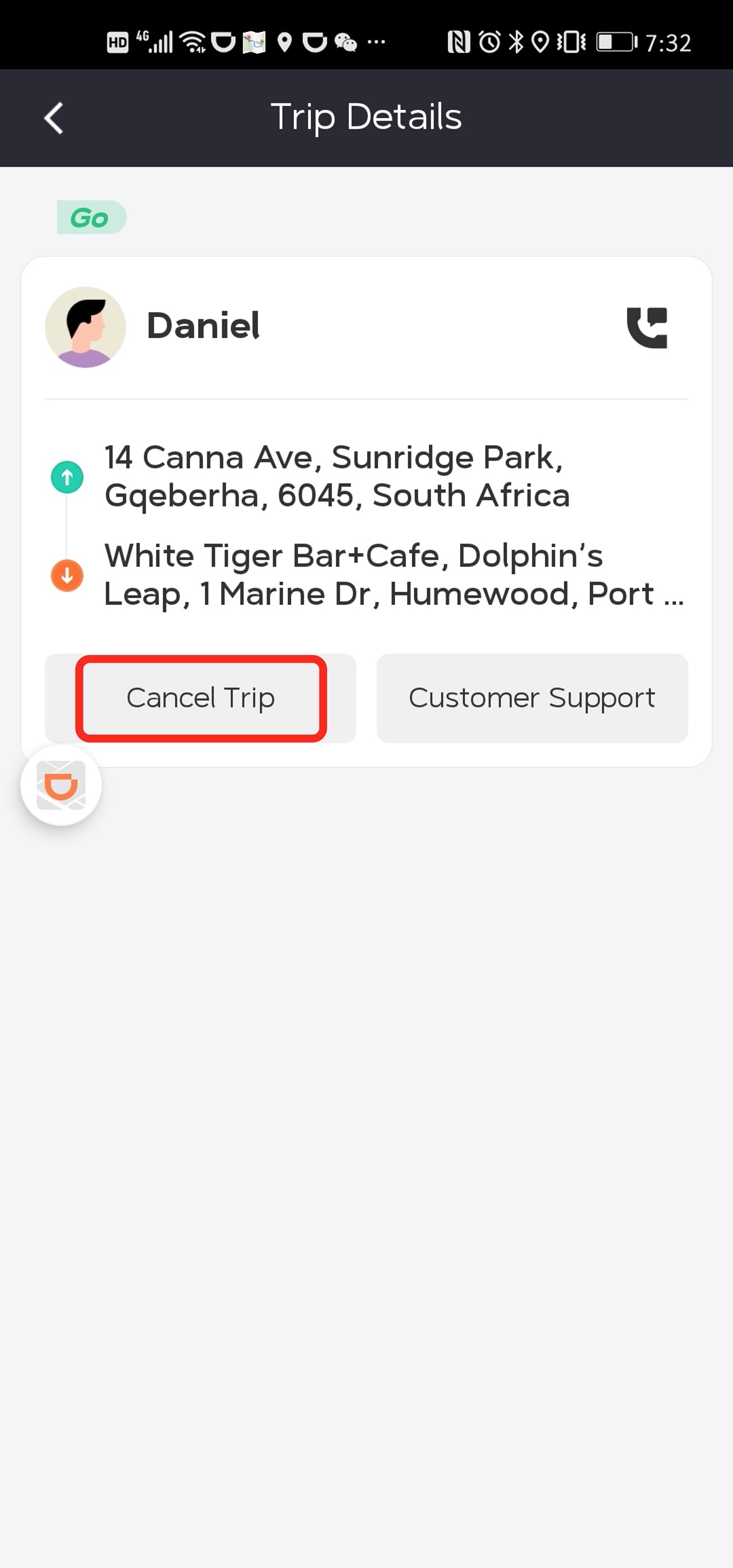}}
	%	\hspace{1ex}
	\subfigure[]{
		\label{fig:intro:sub2}
		\includegraphics[width=0.23\textwidth]{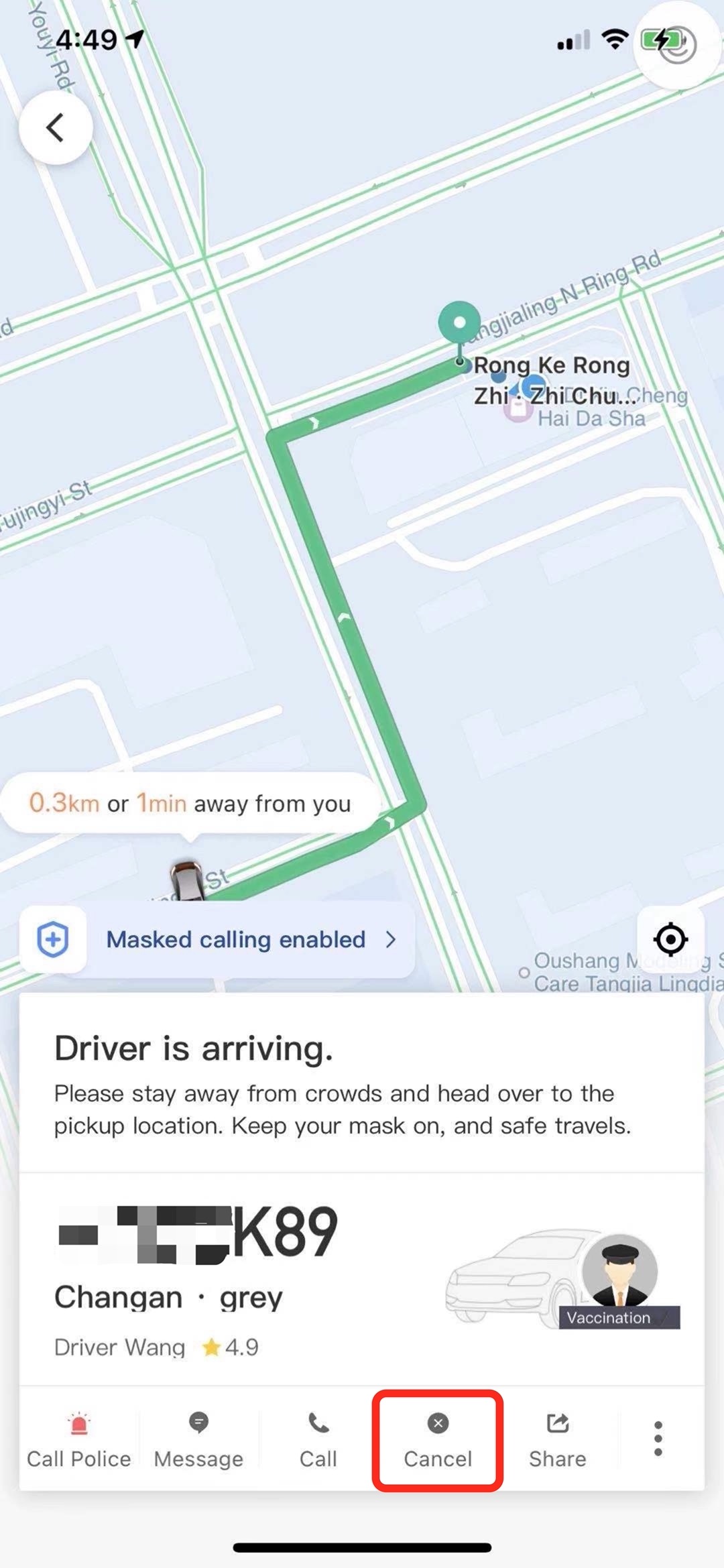}}
	\caption{The operation interfaces for the passenger (a) and the driver (b) after the order was responded by the platform.}
	\label{fig:intro} %% label for entire figure
	%	\vspace{-4ex}
\end{figure}
Specifically, once a passenger makes a request for a ride and an order is generated. Then she/he will be matched to a driver in the surrounding areas according to her/his origin and departure time. Next, as Fig. \ref{fig:intro} shows that both the passenger and driver have the option of cancellation in their user interfaces, and either of them can choose to cancel the matching result by clicking the cancel button. This usually happens for various reasons, e.g., the passenger is concerned about a driver low rating score, or the driver encounters a no-show passenger. It is a common practice for ride-hailing platforms to infer the success rate of a passenger-driver matching pair first, and based on it, an order to the most promising driver can be generated to ensure a smooth service. An accurate prediction can not only improve the real trading volume, but also provide both the passengers and drivers with better user experience.

In this paper, we term the described problem as \underline{\textbf{M}}atching \underline{\textbf{S}}uccess \underline{\textbf{R}}ate (\textbf{MSR}) prediction. Generally, MSR prediction is more challenging compared with traditional advertising problems (e.g., Click-Through Rate (CTR) prediction \cite{qu2016product, huang2013learning}) because it is a collective process which involves two kinds of platform users, i.e., passengers and drivers whose decisions are also highly dependent on the real-time environment. 
MSR represents a bilateral consensus between passengers and drivers and their contexts jointly determine the final results. Hence, MSR prediction is subsumed under a more complicated scenario where we need to fully understand both the passengers' and drivers' intentions as the rapidly changing contexts around them, such as the instant supply-demand ratio and the traffic condition. In other words, it is challenging to capture all the interactions among these complicated factors. Firstly, people's personalities can influence their decisions, e.g., a driver would not show much patience if the passenger is currently late for more than 5 minutes. Secondly, MSR is affected by many external factors. For example, a driver would prefer a passenger who has a longer (i.e., more profitable) trip. A passenger tends to cancel orders matched with drivers who are more than 3 km away, but she/he may not do that if the passenger is in a rural area. Therefore, the sophisticated combinatorial effect among users' preference and the dynamic contexts makes it difficult to learn a precise mapping from these variables to the eventual MSR. 

Furthermore, unlike the online advertising problems that are commonly not limited by the spatial locations, MSR prediction suffers from the severe data imbalance among different cities. Some international metropolises generate millions of orders one day, leading to thousands of orders in one minute, while there are only a few orders during several continuous hours in some small cities. On one hand, models trained for those small cities are subject to inferior performance. On the other hand, compared with fully-developed metropolitan markets, small cities have lots of potential customers that many ride-hailing platforms are contending for. As a result, it is especially crucial yet challenging to provide accurate MSR prediction for small cities whose data are highly scarce.

As MSR prediction is a new-born problem accompanying the recent prosperity of ride-hailing platforms, potential solutions from existing studies are hardly satisfactory. Some studies~\cite{ liu2017functional, zhang2017deep, wang2019unified} on traffic prediction propose deep learning-based models where data imbalance still remains unresolved. In this regards, Yao et al.~\cite{yao2019learning} and Wang et al.~\cite{wang2018crowd} propose meta learning frameworks to transfer knowledge from source cities to target cities for spatiotemporal prediction, but the combinatorial effect among different real-time features are largely overlooked.
Notably, some conceptual similarity can be also found from research on user response prediction \cite{luo2020dynamic, guo2017deepfm, juan2016field}, where a representative task is to predict the possibility that a user will click through a given link/advertisement. Learning feature interactions lies in the core of user response prediction, and methods like Deep Structured Semantic Model (DSSM)~\cite{huang2013learning, shen2014latent, elkahky2015a} and Factorization Machine (FM) based models~\cite{ guo2017deepfm, he2017neural,  lian2018xdeepfm} are capable models for this purpose. 
Recently, to account for the temporal dynamics of features, there are variants adopting Recurrent Neural Network (RNN) \cite{palangi2014semantic, song2016multi} and self-attention networks \cite{chen2020sequence} to fully investigate the sequential dependencies while modelling feature interactions. Unfortunately, user response prediction only considers the decision dynamics from the customer side, while the other side is a display item and considered static. This assumption is obviously ill-posed for MSR prediction which involves a bidirectional, collective decision process. As we have discussed, these methods are not designed to handle the geographically unbalanced data. Moreover, MSR is used to support instant operations like order assignment in a high throughput environment. Considering that the aforementioned prediction methods are usually complex (e.g., the deep residual network in \cite{zhang2017deep} and the convolutional network in \cite{lian2018xdeepfm}),  a lightweight model for MSR prediction is necessary to ensure efficiency for online deployment. 

In this paper, to thoroughly capture feature interactions from both passenger and driver sides, we propose the \underline{\textbf{M}}ulti-\underline{\textbf{V}}iew model (\textbf{MV}) for MSR prediction. In particular, to fully capture the dynamics of the constantly changing context, we build our MV model upon the Differentiable Neural Computer (DNC)~\cite{graves2016hybrid}, a variant of neural memory networks~\cite{graves2014neural} while modelling feature interactions. In addition, with the external memory matrix coupled with DNC, our MV model can effectively accumulate useful knowledge from cities to improve its prediction accuracy. As MSR has to be inferred in real time, we aim to keep the prediction model simple yet powerful for minimizing the latency during online inference. Hence, we further propose a learning scheme based on Knowledge Distillation (KD) \cite{ba2013deep, hinton2015distilling, mishra2017apprentice} to support learning a lightweight model used for real-world deployment. Based on MV, we design a teacher model that utilizes the learned knowledge about data-intensive cities to complement the prediction for the target city. By encouraging the simpler student model to mimic the behavior of the teacher model, the resulted (student) model is fully capable of delivering quality MSR predictions under high data scarcity.

To the best of our knowledge, this is the first work to identify a novel problem that comes along with the prosperity of ride-hailing platforms, namely passenger-driver matching success rate (MSR) prediction. MSR prediction advances existing user response prediction with the notion of bilateral decision-making, and it can generalize to and benefit a wider range of applications like friend-making websites and online marketplaces. Apart from that, the main technical contributions of this work are summarized as follows:
\begin{itemize}
	\item We propose the \underline{\textbf{M}}ulti-\underline{\textbf{V}}iew model (\textbf{MV}) to solve the MSR prediction problem, which provides a comprehensive interaction scheme for features from different perspectives in dynamic contexts, and can retain knowledge about a city for future predictions.
	\item Coupled with MV, we design a \underline{\textbf{K}}nowledge \underline{\textbf{D}}istillation framework (\textbf{KD}) to transfer knowledge from other cities to the lightweight model designed for the target city. This not only mitigates the data imbalance among cities but also results in a compact prediction model for efficient deployment. 
	\item We conduct extensive experiments on real industry-level datasets, where the results demonstrate the superior performance of our approach in both accuracy and scalability.
\end{itemize}

The rest of the paper is organized as follows. 
We first introduce the preliminaries in Section~\ref{sec:preliminaries}. After that, we present the multi-view model in Section~\ref{sec:mv} and knowledge distillation framework in Section~\ref{sec:kd}. We evaluate the proposed solutions in Section~\ref{sec:evaluation}. Then we review related work in Section~\ref{sec:related}. Finally, we conclude this study in Section \ref{sec:conclusion}.

\section{Preliminaries}\label{sec:preliminaries}
%\begin{figure}[htbp]
%	\setlength{\abovecaptionskip}{0.2cm} 
%	\setlength{\belowcaptionskip}{-0.2cm}
%	\centering
%	\includegraphics[width=\linewidth]{./images/OpenStreetMap.pdf}
%	\caption{An example of road network in Beijing}
%%	\Description{beijing by osm}
%\end{figure}

In this section, we define some key concepts and then formulate our research problem. We have also provided a list of notations used in this paper in Table \ref{table:notations}.
\subsection{Definitions}\label{sec:definitions}

\begin{definition}{\textbf{Passenger Request.}}
	Passengers relying on a ride-hailing application have a different way to acquire a ride compared with the traditional roadside hailing. When these passengers need a ride to somewhere, they will make a request denoted as $P_{request}$ $=$ $(ID_p,$ $ts_p,$ $l_r,$ $l_o,$ $l_d,$ $st_p)$ which includes the passenger ID, timestamp, current location, origin and destination locations, status (i.e., responded or not responded, etc). A passenger request is also called an \textbf{order} on the ride-hailing platform.
\end{definition}

\begin{definition}{\textbf{Driver Status.}}
	Every driver has a tuple to describe his/her states $D_{status}$ $=$ $(ID_d,$ $ts_d$, $l_s$, $st_d)$ which consists of driver ID, timestamp, location, and state (i.e., empty, occupied, on the way to a pick-up, etc.).
\end{definition}

\begin{definition}{\textbf{Passenger-Driver Pair.}}\label{def:pdp}
	Once a passenger request is generated, the ride-hailing platform will match a driver for it as soon as possible according to the information provided and the surrounding drivers' status. The passenger-driver pair will be constructed and denoted as $p$ $=$ $(ID_p,$ $ ID_d,$ $ l_o,$ $ l_d,$ $ts)$ which means the passenger $ID_p$ and driver $ID_d$ are paired by the platform at timestamp $ts$. The origin and destination locations of the trip are $ l_o$ and $ l_d$, respectively. Then the passenger and driver will respond to this matching result via acceptance or cancellation. We use $y$ to label this passenger-driver pair. If either of these two participants chooses to cancel this pair then $y$ $=$ $0$, otherwise we have $y$ $=$ $1$ which means they both accept it.
\end{definition}

\subsection{MSR Prediction}\label{sec:msr-definition}
\begin{problem}{\textbf{\underline{M}atching \underline{S}uccess \underline{R}ate (MSR) prediction.}}
	For an arbitrary city, we are given a sequence of $n$ passenger-driver pairs $\mathcal{P}$ $=$ $\{p_1,$ $ p_2,$ $...,$ $ p_i,$ $...,$ $p_{|\mathcal{P}|}\}$ and a corresponding label sequence $\mathcal{Y}$ $=$ $\{y_1,$ $ y_2,$ $...,$ $ y_i,$ $....,$ $y_{|\mathcal{P}|}\}$. $|\mathcal{P}|$ denotes the total number of  matching pairs. \textbf{MSR} prediction is a classification task to predict the match success probability $y_{|\mathcal{P}|+1}$ of any given passenger-driver pair $p_{|\mathcal{P}|+1}$. In MSR prediction, we need to make MSR prediction for a set of \textbf{source cities} $\Phi_s =$ $\{s_1,$ $s_2,$ $...,$ $s_j,$ $...,$ $s_{|\Phi_s|}\}$ with sufficient data, as well as a set of \textbf{target cities} $\Phi_a =$ $\{a_1,$ $a_2,$ $...,$ $a_k,$ $...,$ $a_{|\Phi_a |}\}$ with sparse data (i.e., for an arbitrary source city and target city, their total numbers of passenger-driver pairs within a same time period satisfy $|\mathcal{P}_{a_k}|$$\ll$ $|\mathcal{P}_{s_j}|$). It is worth noting that, when predicting the MSR on target cities, we have an additional goal of improving our model's performance on target cities by borrowing usable knowledge learned from source cities.
\end{problem}

\begin{table}[h!]
	\renewcommand\arraystretch{1.3}
	\centering
	\caption{Key notations used throughout the paper.}
	\label{table:notations}
	\begin{tabular}{m{1.5cm}<{\centering}|m{6cm}<{\centering}}
		\toprule 
		Notation               &Description                                        \\ \hline 
		$P_{request}$, $D_{status}$                         & passenger request and driver status     \\\hline 
		$ID$, $ts$, $ l$, $st$                        & user ID, timestamp, location, status \\\hline 
		$\mathcal{P}$, $\mathcal{Y}$ &a sequence of passenger-driver pairs and labels \\\hline 
		$s_j,$ $a_k,$                    &arbitrary source and target city  \\\hline 
		$\Phi_s,$ $\Phi_a$                   &sets of source and target cities  \\\hline 
		$\textbf{x}_i^{o}$, $\textbf{x}_i^{p}$, $\textbf{x}_i^{d}$                    & features of the order, passenger and driver for the $i$-th passenger-driver pair \\
		\hline 
		$\mathcal{C}_i$                    &  context feature sequence\\
		\hline 
		$\textbf{x}_t^{c}$                    &the context feature at time slot $t$  \\
		\hline 
		$\textbf{e}_i^{\cdot}$ & feature embedding or feature interaction representation \\\hline 
		$\textbf{M}_t$                    & memory matrix at time slot $t$\\
		\hline 
		$\textbf{M}_{s_j},$ $\textbf{M}_{a_k}$                    & memory matrices of arbitrary source city and target city \\\hline 
		$\mathcal{M}_{\Phi_s}$                    &set of source cities' memory matrices  \\\hline 
		$\textbf{r}_{t}^u,$ $\hat{\textbf{e}}_t, $  $\hat{\textbf{a}}_t$                   &read, erase and write vectors at time slot $t$\\
		\hline 
		$\textbf{w}_t^{r, u},$ $\textbf{w}_t^{w},$ & read weight of the $u$-th read operation and write weight at time slot $t$ \\
		\hline 
		$\textbf{h}_t $, $\textbf{E}$, $\bm{\xi}_t $ &hidden state of GRU, unit matrix, and interface vector \\\hline 
		$\textbf{v}^{\cdot}_i$ & final representation of the passenger-driver pair\\\hline 
		$\hat{y}^{\cdot}_i$ & predicted MSR \\\hline  
		\bottomrule
	\end{tabular}	
\end{table}

\section{The Multi-View Model (MV)}\label{sec:mv}
\begin{figure}[th!]
	\centering
	\includegraphics[width=0.48\textwidth]{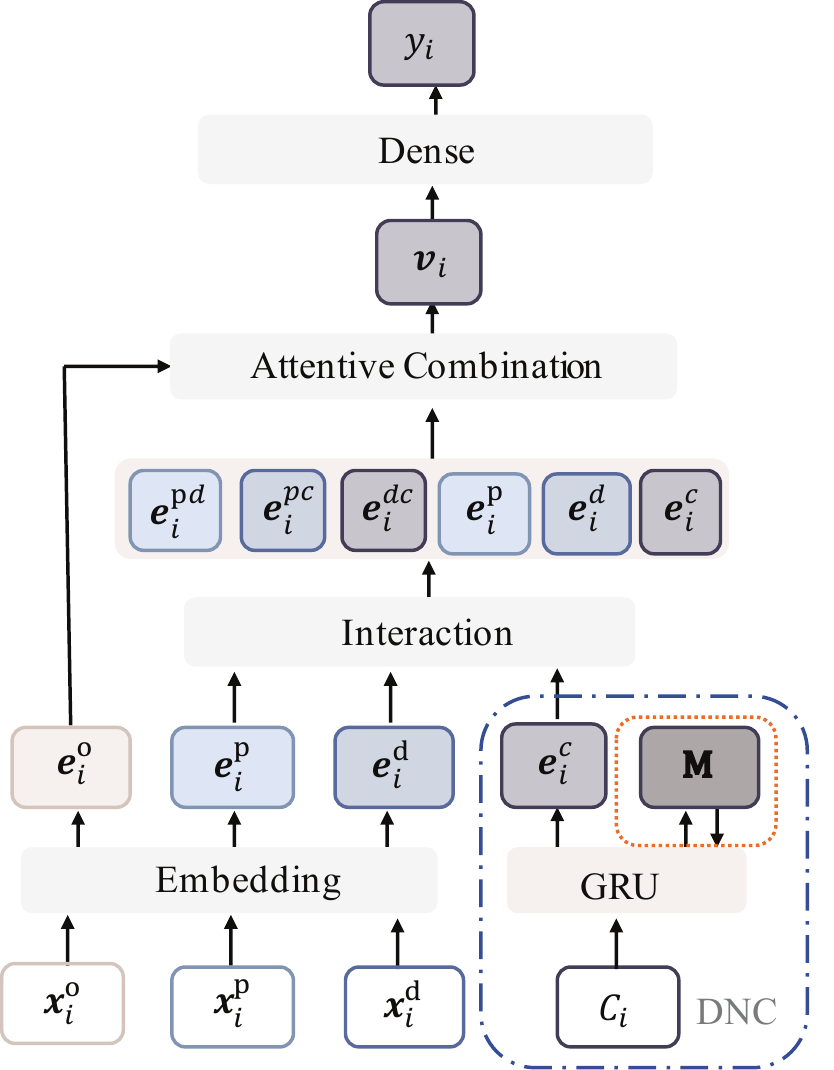}
	\caption{The architecture of the multi-view model. The MV model includes three main steps from the bottom up: feature embedding, multi-view feature interaction modelling, and attentive combination.}
	\label{fig:model} %% label for entire figure
\end{figure}

The collective nature and constantly changing context of MSR require our model to account for the complex interactions of various factors in the real-time. However, unlike CTR prediction models that mainly focus on designing complicated feature interaction paradigms, MSR prediction is also facing severe data imbalance geographically. In light of this, we design a \underline{\textbf{M}}ulti-\underline{\textbf{V}}iew model (\textbf{MV }for short), which not only learns an accurate mapping from the interactions among multiple factors to the MSR in the real time, but also maintains useful knowledge from source cities via memory matrices. Its structure is shown in Fig. \ref{fig:model}. The model consists of four main steps: feature extraction, embedding, interaction and attentive combination. Specifically, we utilize a sequence-based neural memory model for learning the dynamic contexts, which also facilitates the knowledge transfer described in Section \ref{sec:mv}.

\subsection{Feature Extraction and Embedding}
In this section, we category the features into 4 kinds so that we can have a full and interpretable fusion of these features from different perspectives in the following parts.

Given a passenger-driver pair $p_i$, we extract features from $4$ different perspectives: passenger, driver, order and context. The passenger's feature vector $\textbf{x}_i^{p}$ includes the passenger recent behaviors and order history, e.g., the number of his/her canceled orders in the last five minutes, last day and so on. Similarly, the driver's feature vector $\textbf{x}_i^{d}$ consists of the driver's recent behaviors, service types and so on. The order information such as origin, destination and time is important to the MSR of a passenger-driver pair. For instance, a passenger would be more impatient when he/she is in a hurry to work during the morning rush hour. Conversely, passengers would like to wait for a longer time when hanging out on weekends. Hence, the order information is represented as an order feature vector $\textbf{x}_i^{o}$. Last but not least, a context sequence $\mathcal{C}_i$ $=$ $\{\textbf{x}_1^{c},$ $\textbf{x}_2^{c},$ $...,$ $\textbf{x}_t^{c},$ $...,$ $\textbf{x}_{|\mathcal{C}_i|}^{c}\}$ is abstracted for each passenger-driver pair $p_i$, which aims to capture the dynamics of the context during the past few time slots. We will provide more details in Section \ref{sec:evaluation}.

Then, as Fig. \ref{fig:model} shows, we feed $\textbf{x}_i^{o}$, $\textbf{x}_i^{p}$, $\textbf{x}_i^{d}$ into an embedding layer having two steps. Firstly, we project categorical features into embedding vectors and concatenate them with the other continuous and numerical features to reform a complete vector for each perspective. Secondly, we feed each of the three vectors into a dense layer separately and obtain the final embedding vectors (i.e., $\textbf{e}_i^{o}$, $\textbf{e}_i^{p}$, and $\textbf{e}_i^{d}$ ) with a unified embedding size $d$ for subsequent calculations.

\subsection{Context Embedding and Knowledge Building}\label{sec:contextbuilding}
For each city, we can extract a set of contextual features that carry temporal dynamics (e.g., traffic condition) and are of great importance to the MSR. In MV, the transferring knowledge for target cities are built upon the context embeddings of source cities. Recall that we have extracted a sequence $\mathcal{C}_i$ for each passenger-driver pair, which carries the crucial real-time contextual information that heavily impacts the matching results of a passenger-driver pair. In this part, our goal is to learn an effective representation for the dynamic context of the corresponding passenger-driver pair. Firstly, the context representation should encode the influence it would have on the passenger and driver's decisions. Secondly, we aim to obtain a memory of all kinds of context situations as a reference knowledge to help improve future predictions, especially for cities with sparse data. Above all, we design the context embedding component based on DNC \cite{graves2016hybrid} which is extension to Neural Turing Machine (NTM) \cite{graves2014neural}. DNC contains two basic components: a neural network controller and an external memory matrix. The controller interacts with the other model components with its input and output and meanwhile it also reads from and write to the external memory matrix. A sequential controller in DNC could help capture the temporal dynamics in context features and the memory matrix it maintains can be utilized for transferring knowledge from source cities to the target one. 
Besides, DNC's dynamic memory allocation scheme solves the memory overlapping and non-recoverable memory writing problems of NTM, making it a better fit than NTM for our task. We introduce the memory operation and controller network below.

\subsubsection{External Memory.}
The external memory is a fix-sized matrix $\textbf{M}_t$ $\in$ $\mathbb{R}^{n\times m}$. DNC reads from and write to the memory matrix with the help of read/write weights $\textbf{w}_t$ and an interface vector ${\xi}_t$ 
%$=$ $[...;$ $\hat{\textbf{e}}_t;$ $\hat{\textbf{a}}_t;$ $...]$ 
derived from the controller network. DNC selects locations for reading and writing depends via vectors of non-negative numbers whose elements sum to at most 1. The complete set of allowed weights over $n$ locations is the non-negative orthant of $\mathbb{R}^n$ with the unit simplex as a boundary (known as the ``corner of the cube''):
\begin{equation}
	\label{eq:weightset}
	\Delta_n = \{a \in \mathbb{R}^n: a_u \in [0, 1], \sum_{u=1}^{n} a_u \leqslant  1\}.
\end{equation} 

The read operation emits a set of $R$ read vectors \{$\textbf{r}_{t}^1$, $\textbf{r}_{t}^2$, ..., $\textbf{r}_{t}^u$, ..., $\textbf{r}_{t}^R$ \}at each time step $t$. Each read vector is a weighted sum of  $\textbf{M}_t$'s $n$ rows:
\begin{equation}
	\label{eq:readvector}
	\textbf{r}^u_t  = \textbf{M}^{\top}_t \textbf{w}_t^{r, u}, 
\end{equation} 
with $R$ read weights $\textbf{w}_t^{r, 1},$  $\textbf{w}_t^{r, 2},$ ..., $\textbf{w}_t^{r, u},$ ..., $\textbf{w}_t^{r, R}$ $\in$ $\Delta_n$. 
%are used to compute weighted averages of the contents of the locations. 

The write operation is mediated by a single write weight $\textbf{w}_t^{w}$ $\in$ $\Delta_n$, which is used in conjunction with an erase vector $\hat{\textbf{e}}_t$ $\in$ ${[0, 1]}^m$ and a write vector $\hat{\textbf{a}}_t$ $\in$ $\mathbb{R}^m$ (both are emitted by the controller) to modify the memory as follows:
\begin{equation}
	\label{eq:write}
	\textbf{M}_t  = \textbf{M}_{t-1}  \odot (\textbf{E }- \textbf{w}_t^{w}\hat{\textbf{e}}_t^{\top}) + \textbf{w}_t^{w}\hat{\textbf{a}}_t^{\top}, 
\end{equation} 
where $\odot$ denotes element-wise multiplication and $\textbf{E}$ is an $n \times m$ matrix consisting of $1$s. 

\subsubsection{Controller Network.} 
The controller network can be a recurrent or feedforward network but a recurrent controller has its own internal memory that can complement the external memory. Hence, while DNC and NTM use Long and Short Term Memory network (LSTM) \cite{hochreiter1997long}, we adopt GRU \cite{cho2014learning} which performs similar to LSTM but is computationally cheaper \cite{wang2017gated}. As the controller network, GRU would receive the set of read vectors \{$\textbf{r}_{t-1}^1$, ..., $\textbf{r}_{t-1}^R$ \} from the memory matrix $\textbf{M}_{t-1}$. Then the read vectors are concatenated with the current input feature vector $\textbf{x}^c_t$ $\in$  $\mathcal{C}_i$ as the controller's input $\mathcal{X}_t$ $=$ $[\textbf{x}^c_t;$ $\textbf{r}_{t-1}^1$; ...; $\textbf{r}_{t-1}^R]$. For convenience, we formulate the GRU in a simple version:
\begin{equation}
	\label{eq:gru}
	\textbf{h}_t = GRU(\mathcal{X}_t, \textbf{h}_{t-1}),
\end{equation} 
where $\mathcal{X}_t$, $\textbf{h}_t$ are respectively the input and output vector. 
The outputs of controller network are a function of  the hidden states, which can be denoted by the following:
\begin{equation}
	\label{eq:controller1}
	\textbf{e}^c_t = \textbf{W}_y[\textbf{h}^1_t; ...; \textbf{h}^L_t], 
\end{equation}
\begin{equation}
	\label{eq:controller2}
	\bm{\xi}_t = \textbf{W}_{\bm{\xi }}[\textbf{h}^1_t; ...; \textbf{h}^L_t],
\end{equation} 
where $\textbf{e}^c_t$ is the output vector at time slot $t$; $\bm{\xi}_t $ $=$ $[...;$ $\hat{\textbf{e}}_t;$ $\hat{\textbf{a}}_t;$ $...]$ is called the interface vector which is subdivided to parameterize the memory interactions; $\textbf{h}^1_t; ...; \textbf{h}^L_t$ are the hidden states generated by all $L$ GRU layers and $\textbf{W}_{\cdot}$ denotes the trainable weight.

Then with final output vector $\textbf{e}^c_{|\mathcal{C}_i|}$ of the controller network and the current time step's read vectors, the final output $\textbf{e}^c_i$ of the DNC is defined as:
\begin{equation}
	\label{eq:y}
	\textbf{e}^c_i  = \textbf{e}^c_{|\mathcal{C}_i|} + \textbf{W}_r[\textbf{r}_{|\mathcal{C}_i|}^1; ...; \textbf{r}_{|\mathcal{C}_i|}^R],
\end{equation} 
where $\textbf{W}_r$ is a weight matrix that transforms the concatenated read vectors to the same space as $\textbf{e}^c_{|\mathcal{C}_i|}$.

\subsection{Multi-View Feature Interaction and Attentive Combination}\label{sec:interaction}
One of the challenges in predicting MSR is that, the ride-hailing service is a collective process that involves two kinds of platform users i.e., passengers and drivers whose decisions are both highly dependent on the real-time environment. Therefore, we propose a two-step mechanism to thoroughly model and selectively fuse the feature interactions from different views.

The MSR of a passenger-driver pair is influenced by both the passenger and driver's decisions. Meanwhile, the corresponding context and order attributes would affect their final decisions. The same passenger and driver would make different decisions according to different contexts, while under the same context, different passengers and drivers would make different decisions due to different behavioral preferences. 
%Only when both accept the matching, the final result of the matching pair is success, otherwise, it fails. 
As per our discussion, features in different perspectives have interactions with each other, and those features as well as their interactions have different importance to orders under different situations. 
%Inspired by \ac{PNN} \cite{qu2016product, qu2018product} and attention applications \cite{xiao2017attentional}, 
To facilitate the modeling of such information, we propose our own interaction mechanism which includes two steps: (1) the interactions between the embeddings from different perspectives, (2) the attentive combination of the feature representations and their interaction results. 

In the interaction part, we utilize three types of formulation to capture feature interactions, i.e., inner product, element-wise product, and outer product. Then the three types interactions are combined into the final interaction representation. For the features from any two perspectives, their interaction is formulated as follows:
\begin{equation}
	\label{eq:interaction}
	\left.
	\begin{aligned}
		\textbf{e}^{ab}_i  = [I(\textbf{e}^a_i, \textbf{e}^b_i); (\textbf{e}^a_i \odot \textbf{e}^b_i); O(\textbf{e}^a_i, \textbf{e}^b_i)],\\
		ab\in \{pd, pc, dc\},
	\end{aligned}
	\right.
\end{equation} 
where $I( , )$, $O( , )$ represent inner product and outer product, and specially, $O( , )$ includes a summation operation along the second dimension to flatten the resulted matrix from outer product into a vector. 
%As above, $\odot$ denotes the element-wise product. 
%and the details of the other two products is complemented reference to the literature  \cite{qu2018product}. 
After this step, we can receive $\textbf{e}^{pd}_i$, $\textbf{e}^{pc}_i$, $\textbf{e}^{dc}_i$ which carry the interaction contexts by crossing different views.

%except the single-perspective representations, i.e., $\textbf{e}^p_i,$ $\textbf{e}^d_i,$ $\textbf{e}^c_i$, and $\textbf{e}^o_i$.

Feature representations of different perspectives and their interactions have different importances to orders under different situations. Hence, we calculate attentive weights between each one and the order representation $\textbf{e}^o_i$ and their weighted results can be represented as:
\begin{equation}
	\label{eq:attention}
	\left.
	\begin{aligned}
		\textbf{v}^f_i = \varrho(\frac{\textbf{e}^o_{i}\textbf{w}^q_f \cdot(\textbf{e}^f_i\textbf{w}^k_f)^\top}{\sqrt{d}})\cdot\textbf{e}^f_i\textbf{w}^v_f, \\
		f \in \{pd, pc, dc, p, d, c\}
		%		\textbf{e}^{pd}_i, \textbf{e}^{pc}_i, \textbf{e}^{dc}_i, \textbf{e}^{p}_i, \textbf{e}^{d}_i, \textbf{e}^{c}_i
	\end{aligned}
	\right.
\end{equation}
where $\varrho$ represents the element-wise $softmax$ function, $\textbf{w}^k_f$, $\textbf{w}^q_f$, $\textbf{w}^v_f$ are query, key and value weight matrices dedicated to each representation $\textbf{e}^f_i$, $d$ is the dimension of the representation vector, and $\sqrt{d}$ is the scaling factor to smooth the row-wise $softmax$ output and avoid extremely large values of the inner product, especially when the dimension is very high. Then, the concatenation of all weighted results is taken as the final representation of the features in all perspectives:
\begin{equation}
	\label{eq:concat}
	\left.
	\begin{aligned}
		\textbf{v}_i = [\textbf{v}^{pd}_i ; \textbf{v}^{pc}_i ; \textbf{v}^{dc}_i ; \textbf{v}^p_i ; \textbf{v}^d_i ; \textbf{v}^c_i ].
	\end{aligned}
	\right.
\end{equation}
Finally, we employ a dense layer to predict the final result $\hat{y}_i$ of the label $y_i$ which is formulated as:
\begin{equation}
	\label{eq:output}
	\hat{y}_{i} = \sigma(\textbf{v}_i \textbf{w} + b),
\end{equation}
where $\textbf{w}$ and $b$ are trainable parameters in the prediction layer.

\subsection{Optimizing MV}
With the final predicted results $\hat{y}_{i}$ and the label $y_i$, we formulate the loss function as:
\begin{equation}
	\mathcal{L}_{MV} = \mathcal{E}(\hat{y}_{i}, y_i),
\end{equation}
where $\mathcal{E}$ represents the binary cross entropy. All parameters of our model are optimized with the Stochastic Gradient Descent (SGD) method. Specifically, we utilize Adam~\cite{kingma2014adam} which is a widely-used variant of SGD.

\section{Enhancing MV with Knowledge Distillation (KD)}\label{sec:kd}
\begin{figure}[th!]
	\centering
	\setlength{\abovecaptionskip}{0.15cm} 
	\includegraphics[width=0.48\textwidth]{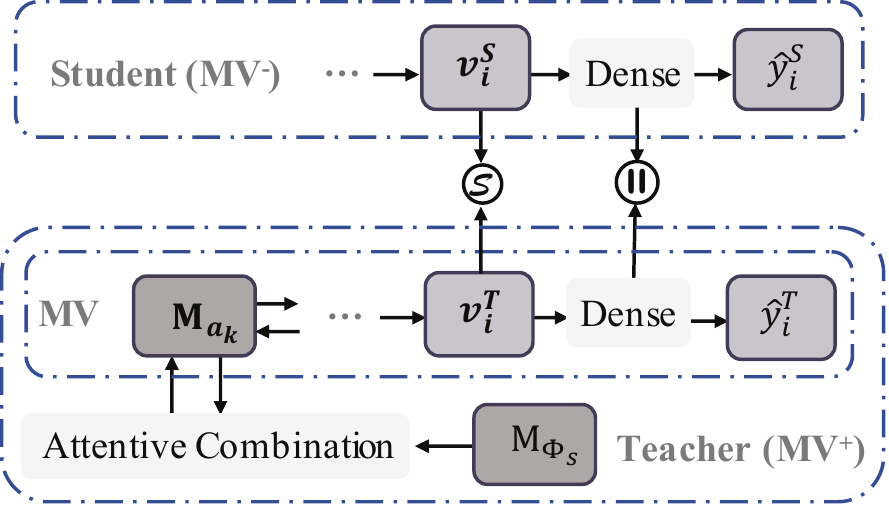}
	\caption{The overview of the knowledge distillation framework. Based on MV, we design a teacher model that utilizes the knowledge learned form data-intensive cities, i.e., $\mathcal{M}_{\Phi_s}$ to complement the prediction for the target city $a_k$, and we also design a simple student model with less parameterization to mimic the behavior of the teacher.}
	\label{fig:framework} %% label for entire figure
\end{figure}

As MSR prediction aims to support online order assignments, it requires real-time efficiency of the predictive model. However, the MV model by itself is computationally heavy for such high throughput applications because of the constant writing and reading operations on the memory matrices. In order to facilitate quicker online inference while ensuring strong performance on cities with high data sparsity, we devise a \underline{\textbf{K}}nowledge \underline{\textbf{D}}istillation framework (\textbf{KD }for short) which has two main components: a complex teacher model and a lightweight student model. The idea is to let the teacher model who can access the information stored in the memory guide the student model, such that the student model being used for online inference can mimic the behaviors of the teacher model and provides high prediction accuracy. As Fig. \ref{fig:framework} shows, it is similar to a Siamese structure where two models are independent except two interactions where $\mathcal{S}$ means to calculate the cosine similarity between the two vectors and $||$ indicates parameter sharing in the dense layer. The framework enables the teacher model to borrow knowledge from source cities with dense data, so that the prediction accuracy can be guaranteed when coping with target cities having fewer records. In addition, by encouraging the student to emulate the prediction behaviors of the teacher with shared parameters, we can obtain a student model that is both lightweight and accurate.
%the simple student model trained under the guidance of  the more complicated teacher model would be used online and improve the efficiency of the online inference. 

\subsection{Memory Preparation}
As the data distributions are diverse among cities, we train a specialized MV model for each city separately. The source cities own sufficient data to facilitate training a quality prediction model, while the data scarcity of target cities tend to lead to poor prediction performance. One of the aims in this section is to solve the problem via knowledge transfer. Firstly, we train the MV models for source cities in $\Phi_{s}$ to obtain the their memory matrices which are denoted as a set $\mathcal{M}_{\Phi_s}$ $=$ $\{\textbf{M}_{s_1},$ $\textbf{M}_{s_2},$ ..., $\textbf{M}_{s_j},$ ..., $\textbf{M}_{|\Phi_s|}\}$. Intuitively, not all source cities' knowledge is of equivalent importance to a target city in $\Phi_{a}$, hence we build the teacher model by attentively select the memorized information from each source city. For a more effective knowledge transfer, we conduct pretraining for the target city as an initialization for its memory matrix $\textbf{M}_{a_k}$ so that the attentive weights between $\textbf{M}_{a_k}$ and each source matrix in $\mathcal{M}_{\Phi_s}$ can be calculated. In this way, a reasonable and effective combination of the memory matrices can be achieved.

\subsection{Setting up Teacher and Student Models}
Fig. \ref{fig:framework} shows the main structures of the teacher and student models, as well as the interactions between them. Note that the teacher and student share the last dense layer, and there is a similarity constraint between their final feature representations, enforcing the student to behave like the powerful teacher model. The teacher model aims to learn as much information as possible so that it can lead to a more accurate result. During the training process of a target city, besides the normal interaction between GRU and the memory matrix, there is a combination procedure. Specifically, the teacher model advances the default MV models by attentively integrating the memory matrices associated with each city and we denote it as $\textbf{MV}^+$ for convenience. Assuming that the target city's memory matrix is $\textbf{M}^i_{a_k}$ w.r.t. sample $p_i$, then $\textbf{M}^{i+1}_{a_k}$ is formulated as:
\begin{align}
	\textbf{M}^{i+1}_{a_k} = \textbf{M}^{i}_{a_k} + \sum_{j=1}^{|\mathcal{C}_s|} \varphi(\frac{\textbf{M}^i_{a_k}\textbf{W}^q_{a_k} \cdot(\textbf{M}_{s_j}\textbf{W}^k_{a_k})^\top}{\sqrt{m}})\cdot\textbf{M}_{s_j}\textbf{W}^v_{a_k}, 
\end{align}
where $\textbf{W}^k_{a_k}$, $\textbf{W}^q_{a_k}$, $\textbf{W}^v_{a_k}$ are query, key and value weight matrices dedicated to each target city, $\varphi$ denotes the row-wise $softmax$ function and $m$ is the column dimension of the memory matrix. 

To this end, it is obvious that the teacher model is heavier than MV in parameters (i.e., the teacher has the additional query, key and value weight matrices compared with MV) which further hinders its practicality for online deployment. Therefore, we devise the student model by stripping away the heavy memory interaction part as Fig. \ref{fig:model} shows with the rectangle in orange and round dot line (i.e., the student omits the Eq.(\ref{eq:controller2}) and Eq.(\ref{eq:y}) compared with MV and their corresponding weights are $\textbf{W}_{\bm{\xi}}$ and $\textbf{W}_{r}$) and we denote it as $\textbf{MV}^-$ for convenience. Consequently, the space complexity of student is reduced by the memory interaction weights $\textbf{W}_{\bm{\xi}}$, $\textbf{W}_{r}$ and the memory aggregation weights $\textbf{W}^k_{a_k}$, $\textbf{W}^q_{a_k}$, $\textbf{W}^v_{a_k}$ compared with the teacher, which provides minimal parameterization and efficient inference in return. It is worth noting that, our model design also facilitates multi-thread processing in the production environment, where each city is deployed at a single computation node and the implementation can easily scale up to a large number of cities.

\subsection{Optimizing KD}
As we can see from Fig. \ref{fig:framework}, the final outputs of  teacher and student models are $\hat{y}^T_i$ and $\hat{y}^S_i$, respectively. Meanwhile, both of them generate a representation vector of all features for each passenger-driver pair $p_i$, i.e., $\textbf{v}^T_i$ and $\textbf{v}^S_i$. The rationale of the knowledge distillation is to make the output of teacher and student model as similar as possible so we formulate the loss function of the teacher-student framework as following:
\begin{equation}
	\mathcal{L}_{KD} = \alpha\mathcal{E}(\hat{y}^T_i, y_i) + \beta\mathcal{E}(\hat{y}^S_i, y_i) - \gamma cos(\textbf{v}^T_i, \textbf{v}^S_i),
\end{equation}
where $cos$ denotes the cosine similarity between two vectors; $\alpha$, $\beta$, and $\gamma$ refer to weighting factors to prioritize the output of a certain loss function over the other, which are learnable in our model.
Similar to MV, all parameters of the knowledge distillation framework are optimized with the variant of SGD, i.e., Adam~\cite{kingma2014adam}.

\section{Evaluation}\label{sec:evaluation}
In this section, we conduct experiments on industry-level datasets to showcase the effectiveness of our model on MSR prediction. In particular, we aim to answer the following research questions via experiments:
\begin{itemize}
	\item[\textbf{RQ1:}] Can MV outperform strong baselines on MSR prediction?
	\item[\textbf{RQ2:}] How does MV benefit from each component of the proposed model structure?
	\item[\textbf{RQ3:}] How does each key hyperparameter affect the performance of MV?
	\item[\textbf{RQ4:}] Is the lightweight model learned via KD effective in predicting MSR under high data sparsity?
	\item[\textbf{RQ5:}] How does each key hyperparameter affect the performance of KD?
	\item[\textbf{RQ6:}] Is our solution to MSR prediction scalable?
\end{itemize}

\begin{table}[h!]
	\setlength{\abovecaptionskip}{0.1cm}
	\renewcommand{\arraystretch}{1.3}
	\centering
	\caption{A summary of datasets in use. \#N denotes the percentage of negative samples among instances.}
	\label{table:datasets}
	%	\vspace{0.02mm}
	%\small
	\begin{tabular}{c c c c c}
		\toprule 
		City     &\#Instance   &\#N  &Size ($km^2$) &Population                  \\ \hline 
		BJ  &31,062,865 &11.45\%&16,411 &21,536,000\\
		SH &21,132,759 & 14.41\%&6,340.5&24,281,400\\
		SZ &12,792,586 &12.16\%&1,997.47 &13,438,800\\
		CD &12,432,933&11.27\% &14,335&16,581,000\\
		DZ &477,156&11.04\%&10,356 &5,748,500\\
		ZZ &858,138 & 7.93\%&12,600&5,160,000\\
		\bottomrule
	\end{tabular}
\end{table}

\begin{figure*}[tb!]
	\centering
	\setlength{\abovecaptionskip}{0.10cm}
	\subfigure[]{
		\label{fig:eva:sub1}
		\includegraphics[width=0.3\textwidth]{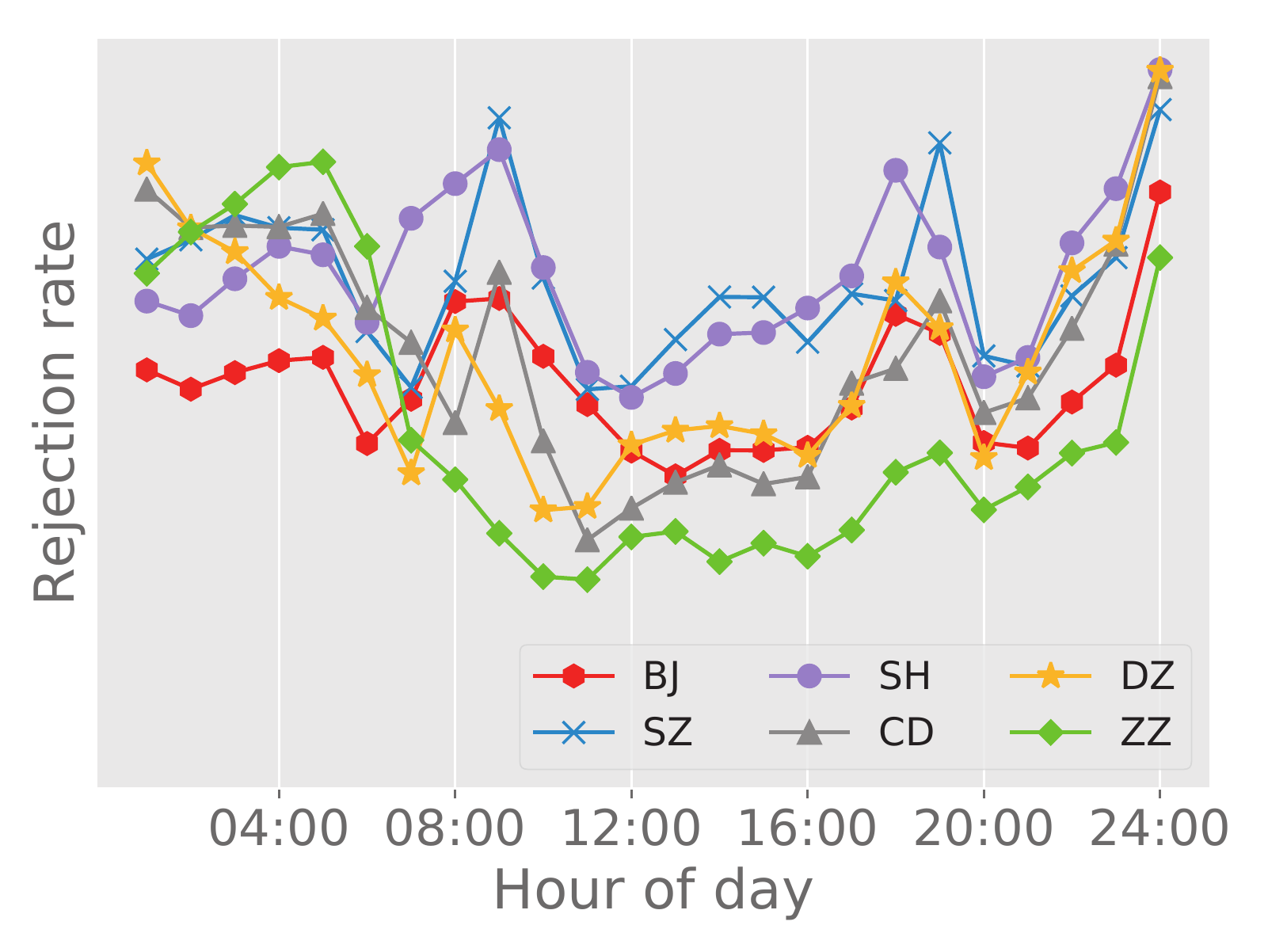}}
	\hspace{1ex}
	\subfigure[]{
		\label{fig:eva:sub2}
		\includegraphics[width=0.3\textwidth]{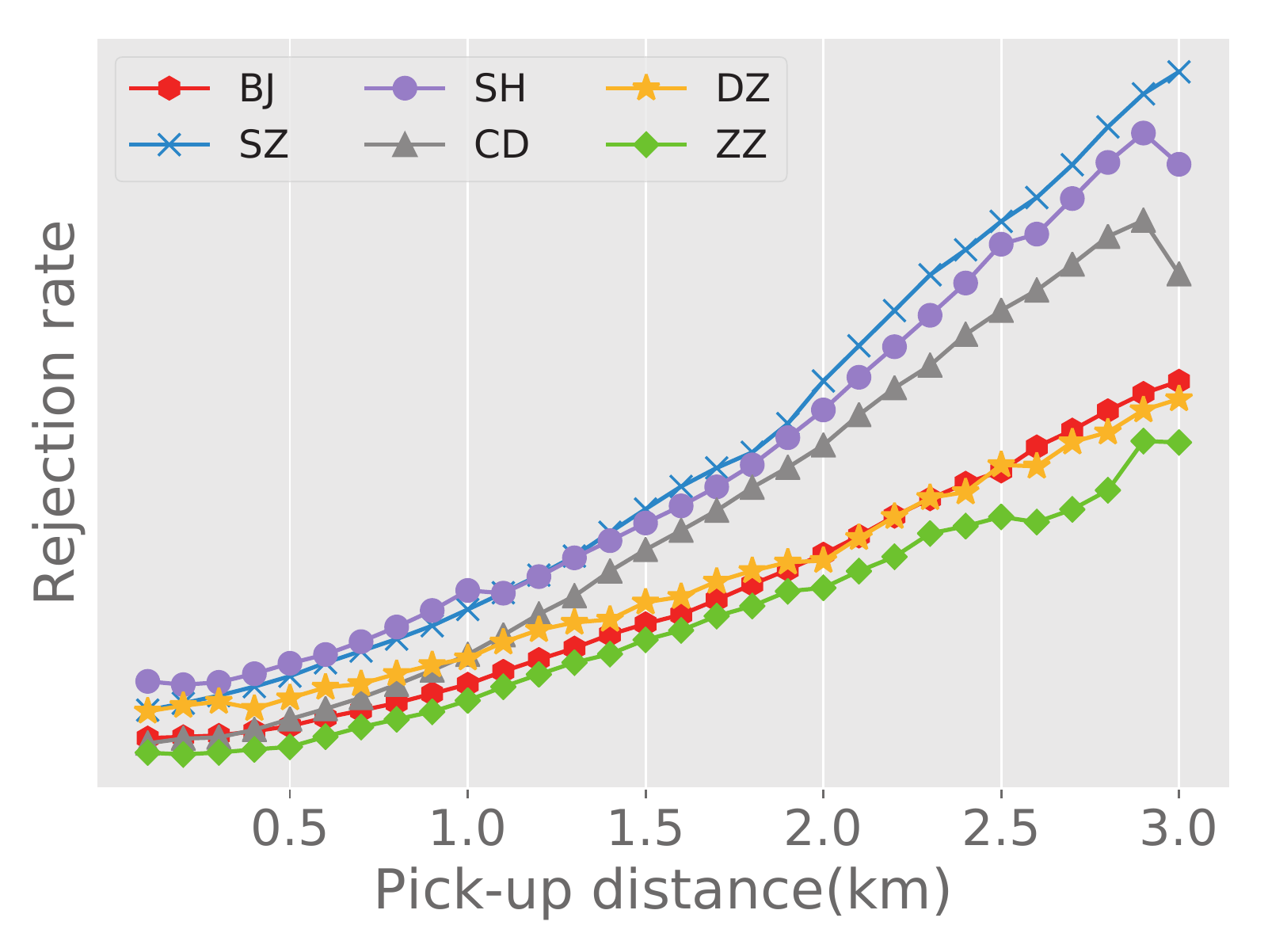}}
	\hspace{1ex}
	\subfigure[]{
		\label{fig:eva:sub3}
		\includegraphics[width=0.3\textwidth]{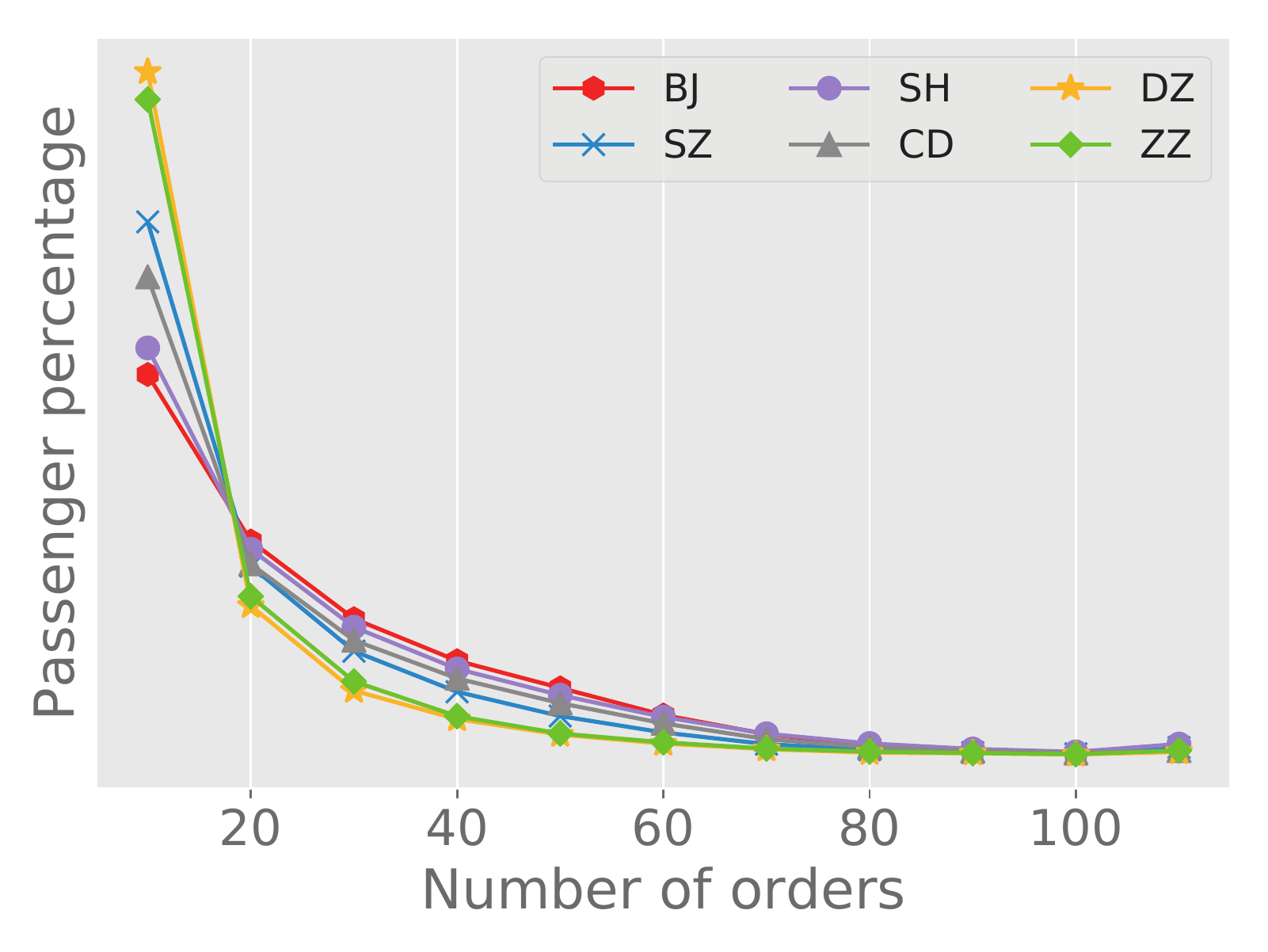}}
	\caption{Statistics of the order rejection rate and the distribution of passengers with different orders. Note that due to security regulations from Didi Chuxing, we are only permitted to showcase the trends within different datasets instead of explicit numbers. }
	\label{fig:eva} %% label for entire figure
\end{figure*}

\begin{figure*}[t]
	\centering
	\setlength{\abovecaptionskip}{0.10cm}
	\subfigure[Beijing]{
		\label{fig:eva0:sub1}
		\includegraphics[width=0.23\textwidth]{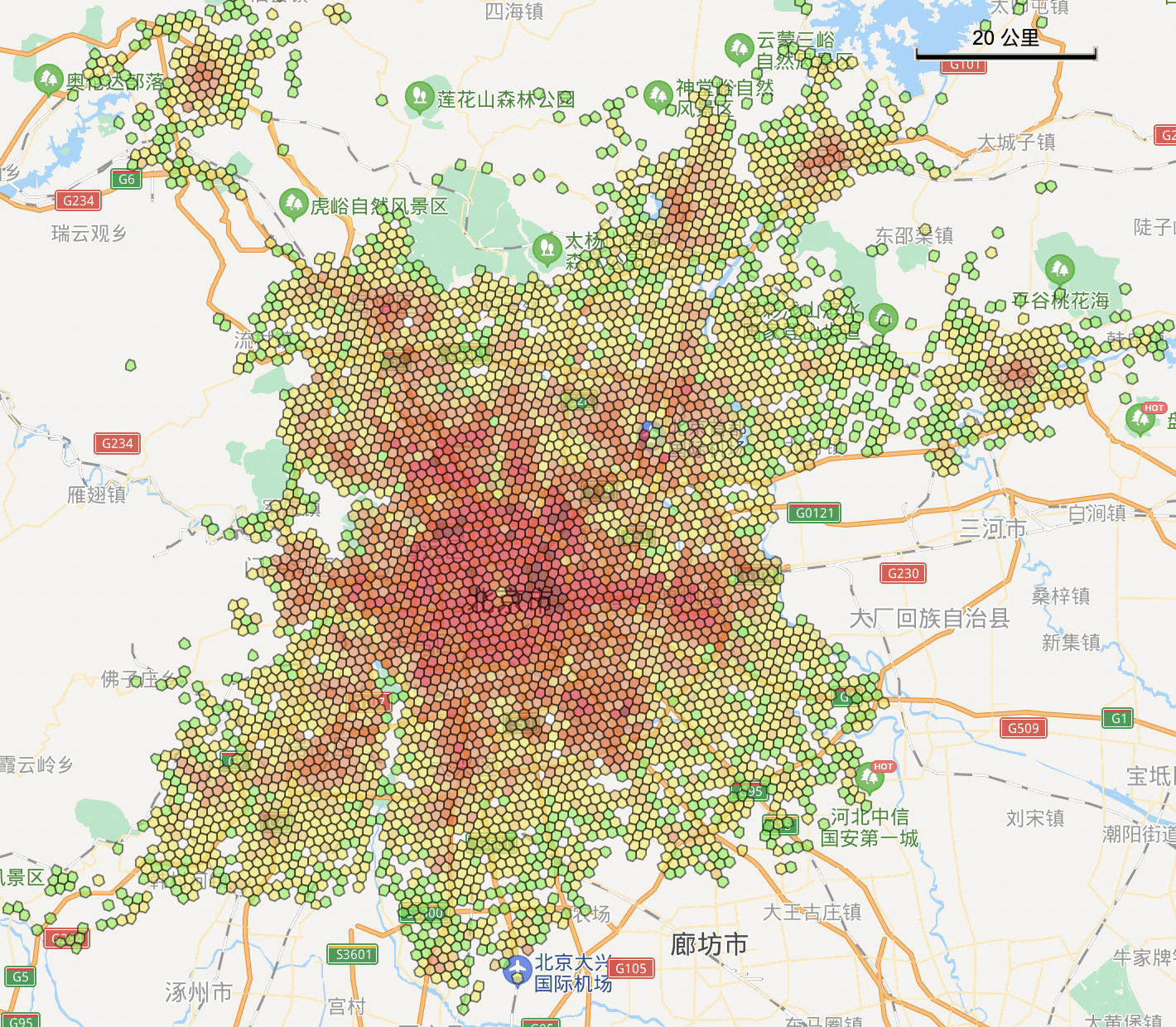}}
	\hspace{1ex}
	\subfigure[Shanghai]{
		\label{fig:eva0:sub2}
		\includegraphics[width=0.23\textwidth]{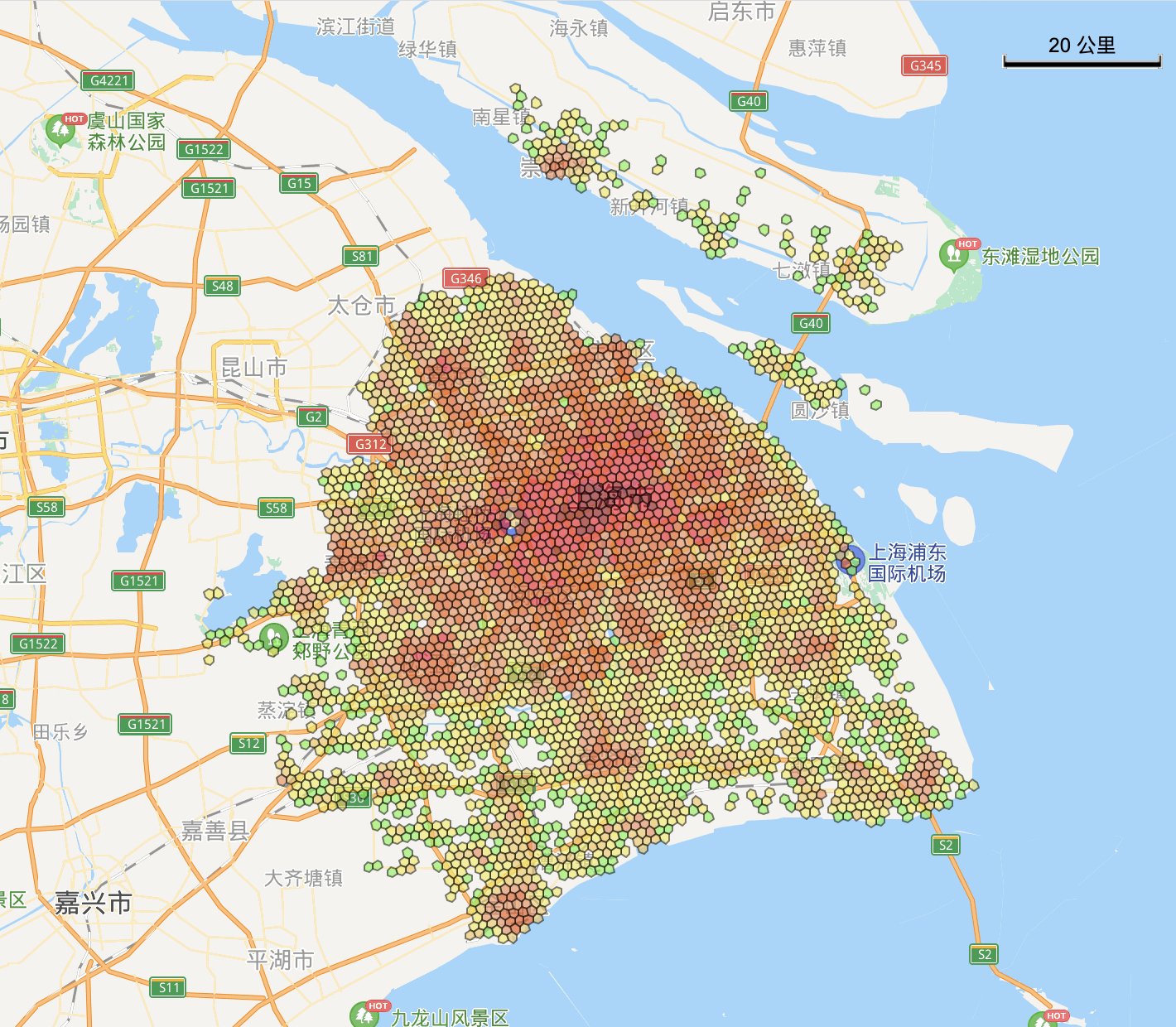}}
	\hspace{1ex}
	\subfigure[Dezhou]{
		\label{fig:eva0:sub3}
		\includegraphics[width=0.23\textwidth]{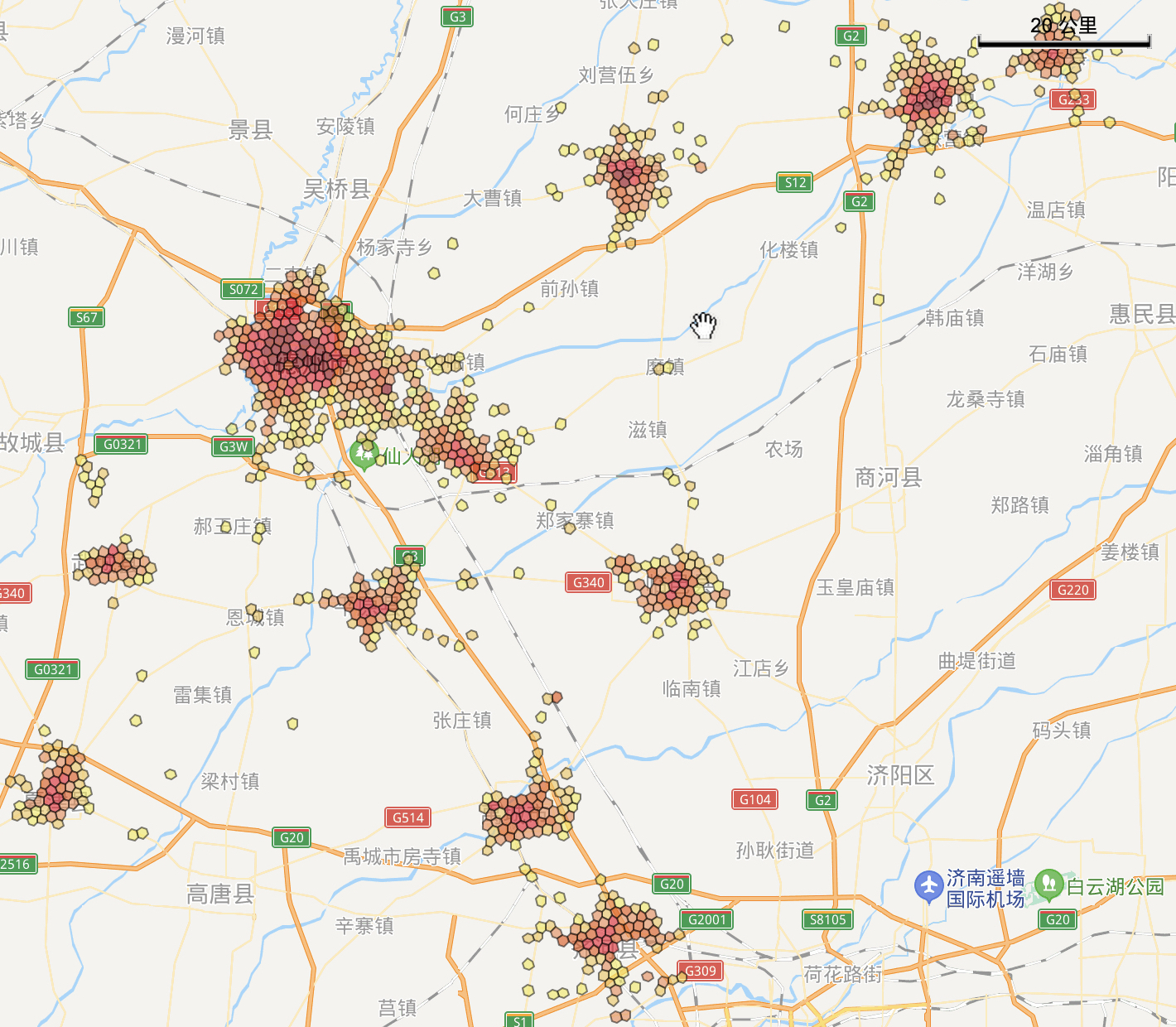}}
	\hspace{1ex}
	\subfigure[Zhangzhou]{
		\label{fig:eva0:sub4}
		\includegraphics[width=0.23\textwidth]{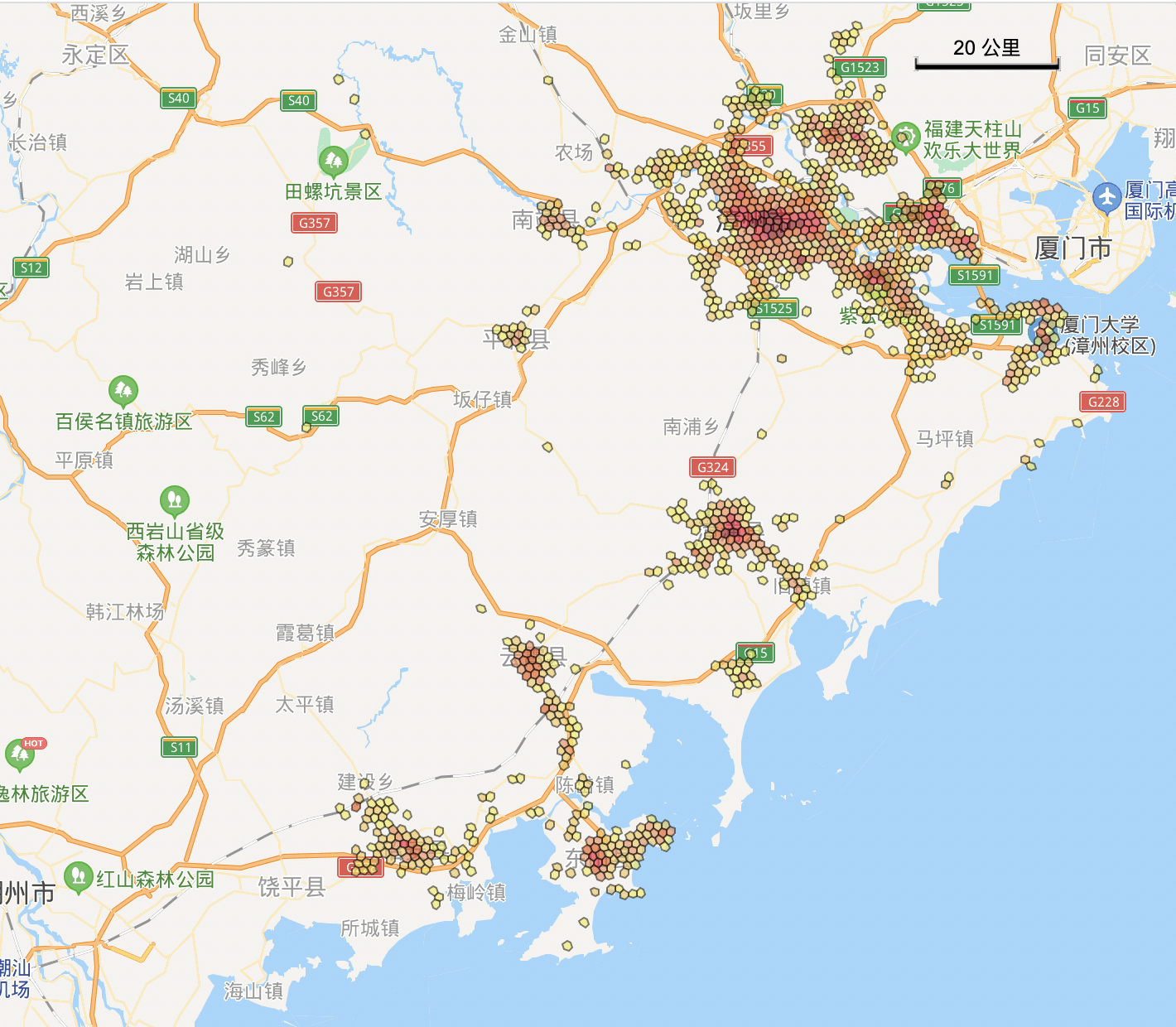}}
	%	\vspace{-2ex}
	\caption{Heat maps of order rejection rates in different areas of different cities, where a darker color represents a higher rejection rate. Exact numbers are also omitted due to security regulations. }
	\label{fig:eva0} %% label for entire figure
	%	\vspace{-4ex}
\end{figure*}

\begin{table}[h!]
	\renewcommand\arraystretch{1.3}
	\centering
	\caption{Features extracted from our datasets.}
	\label{table:features}
	\begin{tabular}{m{1cm}<{\centering}|m{6cm}<{\centering}}
		\toprule 
		Feature & Description \\ \hline 
		\multirow{2}{*}{$\textbf{x}_i^{p}$, $\textbf{x}_i^{d}$}  
		& the number of different historical events during last day/week/month \\\cline {2-2}
		& the number of finished trips during last day/week/month \\\hline 
		\multirow{3}{*}{$\textbf{x}_i^{o}$}  
		& start POI type \\\cline{2-2}
		& end POI type\\\cline {2-2}
		& product type \\\hline 
		\multirow{7}{*}{$\textbf{x}_t^{c} \in \mathcal{C}_i$}  
		& estimated pick-up distance\\\cline{2-2}
		& estimated pick-up time\\\cline {2-2}
		& whether it is a peak period\\\cline {2-2}
		& whether it  is a holiday\\\cline {2-2}
		& whether it is a hot spot\\\cline {2-2}
		& real-time supply and demand ratio\\\cline {2-2}
		&the cancellation rate during the last ten minutes at the starting point\\\hline 
		\bottomrule
	\end{tabular}	
\end{table}

\subsection{Datasets}
We conduct experiments on real-world datasets generated by Didi Chuxing\footnote{https://www.didiglobal.com/}, the largest Chinese ride-hailing platform. We use data collected from 6 cities, namely, Beijing (BJ), Shenzhen (SZ), Shanghai (SH),  Chengdu (CD), Dezhou (DZ), and Zhangzhou (ZZ). Table \ref{table:datasets} summarizes the statistics of these datasets, which are randomly drawn from the platform's daily operations in December 2020. Note that all samples in our datasets are matched orders (i.e., passenger-driver pairs in Definition \ref{def:pdp}), where we label the rejected ones as negative and fulfilled ones as positive, respectively. 

To further showcase the characteristics of our experimental datasets, we provide some statistical visualizations in Fig. \ref{fig:eva} and Fig. \ref{fig:eva0}. From Fig. \ref{fig:eva:sub1} we can see three obvious peaks at about 9:00, 18:00 and 24:00 in five cities. During those three major commuting time periods, people tends to have less patience waiting for a delayed ride, so the rejection rates are high. The data of Zhangzhou presents a different curve, whose peaks appear much earlier morning, which may reflect a specific regular routine of people in this city. As Fig. \ref{fig:eva:sub2} depicts, it is within our expectation that the order rejection rate rises as the pick-up distance increases. However, when the pick-up distance is increased to a larger value, the rejection rates in three cities significantly decreases. This is reasonable because people are more willing to wait when available nearby vehicles are in short. Fig. \ref{fig:eva0} shows heat maps of order rejection rates in different areas of different cities. The closer the distance between an area to the downtown, the higher order rejection rates appear. As implied by the visualizations, there is almost no consistent patterns across all cities while various factors have complex combinatorial impact on MSR, which greatly motivates our model design.

Based on the scale of available instances, we use BJ, SZ, SH and CD as the \textbf{source cities} and DZ, ZZ as the \textbf{target cities} to facilitate knowledge distillation for MSR prediction under data scarcity. Referring to the data sparsity, we conduct a further analysis buy using Fig.4 (c) to illustrate the distribution of passengers w.r.t. the amount of orders placed. It can be seen that target cities have substantially more passengers with fewer orders, and vice versa. Combining the numbers of instances in different cities (Table 2) with the curve in Fig.4 (c), it implies severe data sparsity in target cities caused by inactive passengers, which is also tackled by our model design.
Table \ref{table:features} lists the features we have extracted from four different views (i.e., passenger, driver, order and context) that are used as the input of our model.

\subsection{Baselines}
To evaluate the performance of our model and framework, we make comparison two types of baselines: feature interaction-based methods and knowledge transfer-based methods. %and variants of our proposed MV model.
First, we compare MV with \textbf{feature interaction-based methods} that are designed for user response prediction in the following:
\begin{itemize}
	\item{\textbf{DeepFM:}} DeepFM~\cite{guo2017deepfm} is an end-to-end deep model that emphasizes both low- and high-order feature interactions, which combines the power of factorization machines and neural networks for interaction modelling and representation learning.
	\item{\textbf{PNN:}} The Product-based Neural Network (PNN) ~\cite{qu2016product} utilizes an embedding layer to learn a distributed representation of the categorical data, a product layer to capture inter-field patterns, and fully connected layers to explore high-order feature interactions.
	\item{\textbf{DIN:}} The Deep Interest Network (DIN)~\cite{zhou2018deep} is developed and deployed in the display advertising system in Alibaba. DIN represents users’ diverse interests with an interest distribution and designs an attention network structure to selectively activate the related user interests.
	\item{\textbf{DCN:}} The Deep and Cross Network (DCN)~\cite{wang2017deep} keeps the benefits of a DNN model, and beyond that, it introduces a novel cross network that is more efficient in learning certain bounded-degree feature interactions. In particular, DCN explicitly applies feature crossing at each layer, requires no manual feature engineering, and adds negligible extra complexity to the DNN model.
	\item{\textbf{DSSM:}} The Deep Structured Semantic Model (DSSM)~\cite{huang2013learning} learns representations for both queries and documents in a common low-dimensional space, and then computes their similarity to infer the probability of a matching pair.
	\item{\textbf{Wide\&Deep:}} The Wide\&Deep~\cite{cheng2016wide} model jointly trains wide linear models and deep neural networks to combine the benefits of memorization and generalization for user response prediction. 
	\item{\textbf{DeepCrossing:}} DeepCrossing~\cite{shan2016deep} stacks multiple residual units upon the concatenation layer of feature embeddings in order to learn deeper cross interactions of features.
\end{itemize}
We further compare our proposed KD method in their capabilities of utilizing knowledge transfer to enhance prediction performance on target cities. As for \textbf{knowledge transfer-based methods}, \cite{wang2018crowd, yao2019learning} can transfer knowledge between cities but their prediction targets focus on spatiotemporal time series data, which is unsuitable for our MSR prediction problem. Hence, following~\cite{yao2019learning}, we conduct experiments by designing two fine-tuning strategies for MV as our knowledge transfer-based peer methods: 
\begin{itemize}
	\item\textbf{Single-FT:} We train the MV model on every source city and fine-tune the model for all target cities, respectively. 
	\item\textbf{Multi-FT:} We train the MV model on randomly chosen samples from all source cities and then fine-tune it for all target cities.
\end{itemize}

\subsection{Experimental Settings}
As our problem is a classification task, we evaluate the prediction accuracy with the widely-applied metric, Area Under roc Curve (AUC) and Root Mean Squared Error (RMSE) \cite{chen2020sequence}. In experiments, the ratio between the test set and training set is 1:9. In the training set, we further hold out its last 10\% of the data for validation. We implement our solutions with
TensorFlow 1.15 and Python 3.6. The default sizes of embeddings and the memory matrix are set as $64$ and $16 \times 16$, respectively. The batch size is set to $256$. 
%{\color{red}We have fully tuned all hyperparameters of all baselines among the values that their authors recommend in papers with the Grid Search method. Hence, the results are their best performance.}

\begin{table*}[t]
	\renewcommand\arraystretch{1.3}
	\setlength{\abovecaptionskip}{0.1cm}
	\setlength\tabcolsep{6pt}
	\centering
	\caption{Results of different methods on both source and target cities compared with MV.}
	\begin{tabular}{|c|c|c|c|c|c|c|c|c||c|c|c|c|}
		\hline
		\multirow{2}*{Method}&\multicolumn{2}{c|}{BJ}&\multicolumn{2}{c|}{SZ}&\multicolumn{2}{c|}{SH}&\multicolumn{2}{c||}{CD}&\multicolumn{2}{c|}{DZ}&\multicolumn{2}{c|}{ZZ}\\ \cline{2-13}
		&AUC&RMSE&AUC&RMSE&AUC&RMSE&AUC&RMSE&AUC&RMSE&AUC&RMSE\\ \hline
		DeepFM& 0.7089   & 0.3122     & 0.7116    & 0.3326   & 0.7379   & 0.3385   & 0.7245   & 0.2761  & 0.6834  & 0.3013    & 0.6925  & 0.2658   \\
		PNN& 0.7283  & 0.3074     & 0.7343   & 0.3295    &0.7785   & 0.3285    & 0.7519   & 0.2726 &0.6886  & 0.2984    & 0.7036    & 0.2657    \\		
		DIN& 0.7291   & 0.3048     &0.7253    & 0.3348    & 0.7617    &0.3344   & 0.7462  & 0.2742 & 0.6981  & 0.2988    & 0.7072   & 0.2638   \\
		DCN& 0.7345   & 0.3079    & 0.7327    & 0.3275   &0.7654   & 0.3307    &0.7457  & 0.2721 & 0.6922   & 0.3001    & 0.7036   & 0.2629   \\
		DSSM& 0.7376  & 0.3048    &0.7347   & 0.3305    & 0.7743   & 0.3284    & 0.7464   & 0.2731  &0.6967  & 0.3003   & 0.713   & 0.2649    \\
		Wide\&Deep& 0.7423   & 0.3056    & 0.736    & 0.3277   & 0.7791   & 0.3280   & 0.7469   & 0.2718 & 0.7009   & 0.2975    & 0.7121   & 0.2621      \\
		DeepCrossing& 0.7487   & 0.3043    & 0.7409   & 0.3285   & 0.7848    & 0.3226    & 0.7536  & 0.2731  & 0.6919  & 0.2992    & 0.7063   & 0.2614    \\
		\textbf{MV}& \textbf{0.7562}	& \textbf{0.3012}	& \textbf{0.7538}	& \textbf{0.3228}	& \textbf{0.7998}	& \textbf{0.3201}	&\textbf{ 0.7650	}& \textbf{0.2665}	& \textbf{0.7045	}& \textbf{0.2961}	& \textbf{0.7212}	& \textbf{0.2590  } \\
		\hline
		\hline
		MV-S1&0.7454 &0.3079 &0.7458 &0.3269 &0.7945 &0.3223	&0.7583	&0.2683	&0.6899	&0.2994	&0.7098	&0.2614 \\ 
		MV-S2 &0.7458	&0.3075	&0.7466	&0.3274	&0.7933	&0.3237	&0.7574	&0.2705	&0.6886	&0.3001	&0.7079	&0.2600 \\ 
		MV-S3& 0.7449	&0.3073	&0.7326	&0.3307	&0.7939	&0.3221	&0.7601	&0.2691&0.6859	&0.2998	 &0.7104	 &0.2592 \\ 
		MV-S4& 0.7524	&0.3026 &0.7520	&0.3239	&0.794	&0.3219	&0.7639	&0.2695	&0.6877	&0.2994	&0.7063	&0.2622  \\  
		\hline
	\end{tabular}
	\label{table:overall-mv}
\end{table*}

\begin{figure*}[t]
	\centering
	\setlength{\abovecaptionskip}{0.10cm}
	\subfigure[]{
		\label{fig:eva1:sub1}
		\includegraphics[width=0.23\textwidth]{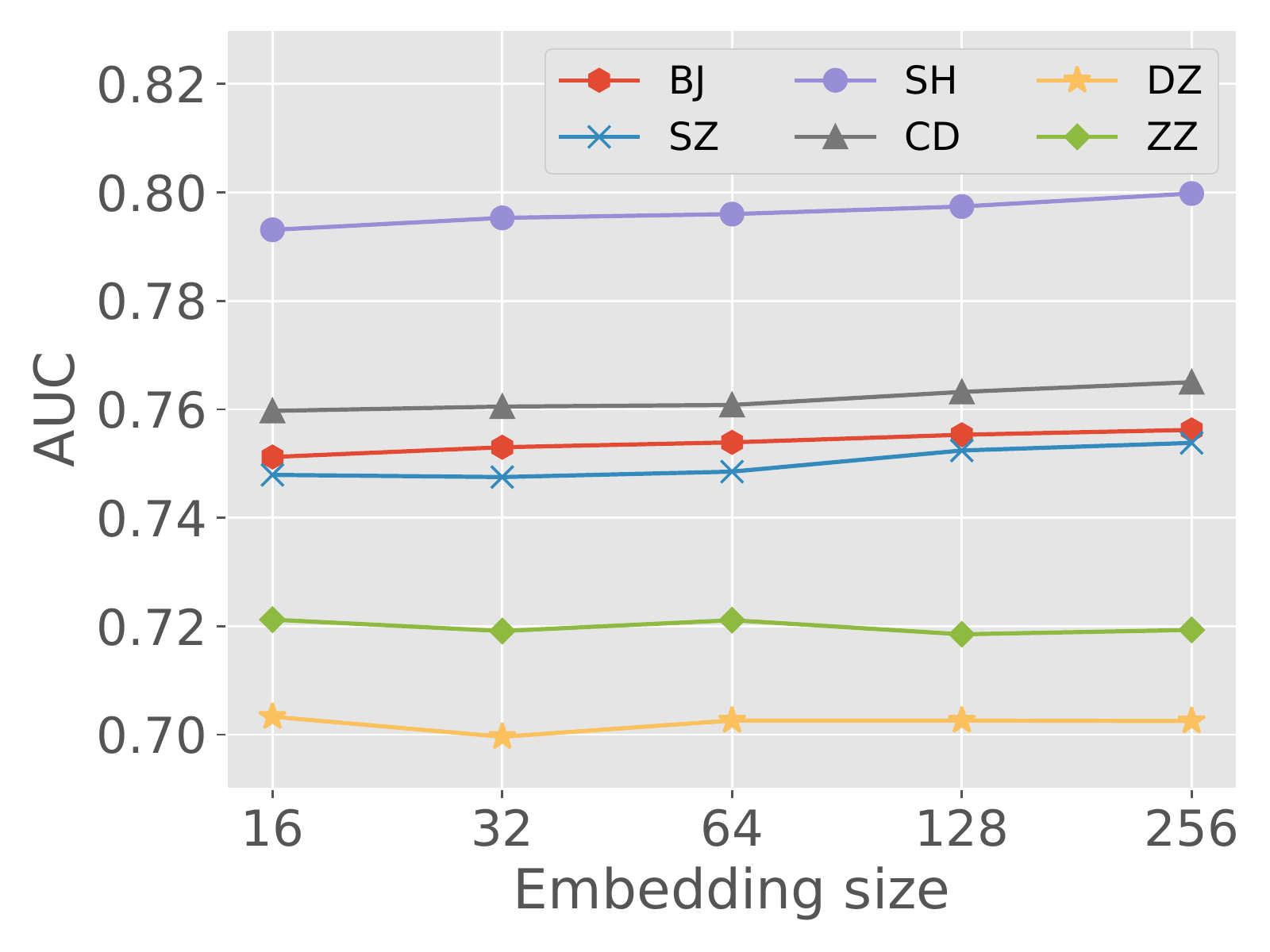}}
	\hspace{1ex}
	\subfigure[]{
		\label{fig:eva1:sub2}
		\includegraphics[width=0.23\textwidth]{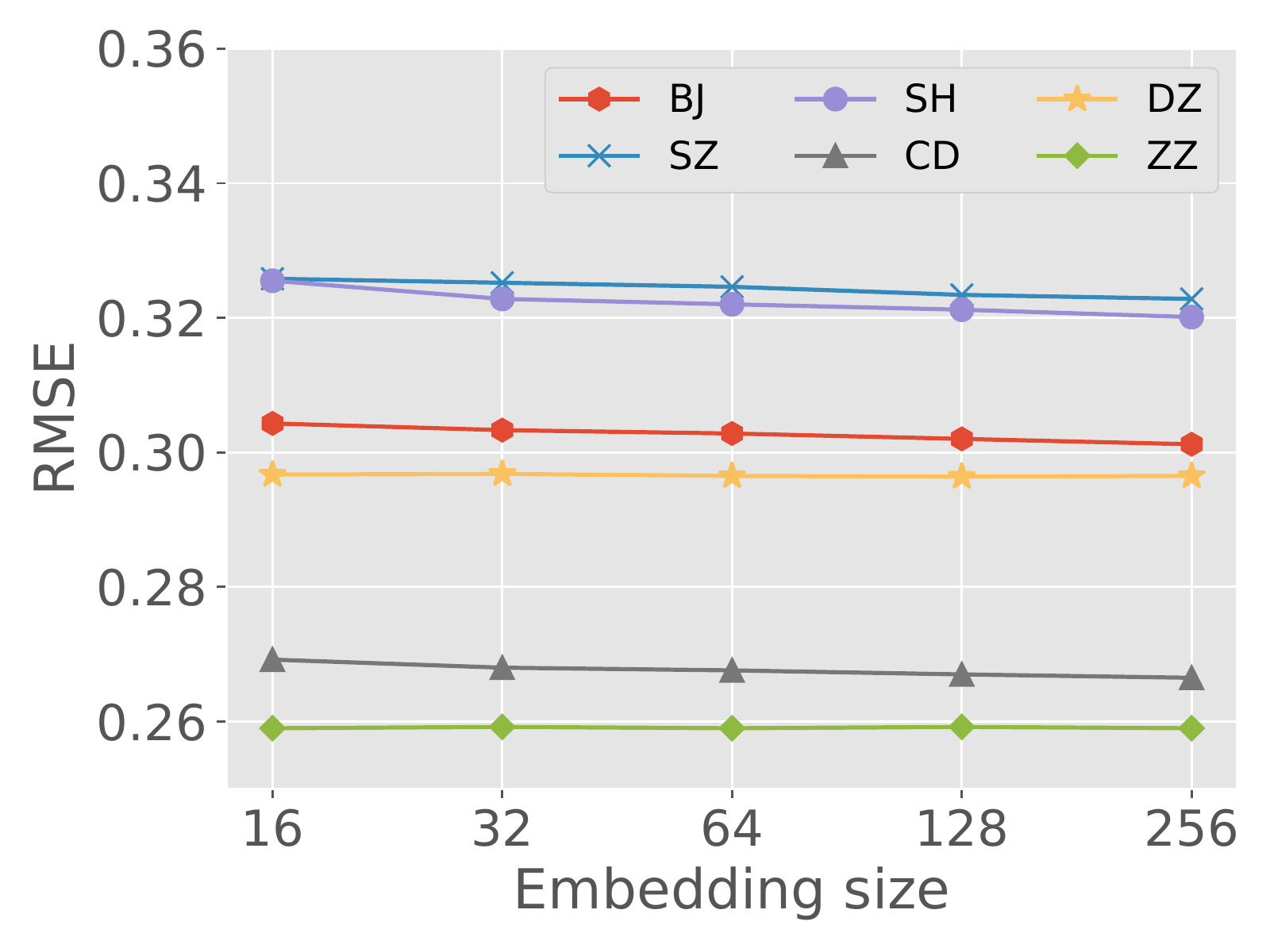}}
	\hspace{1ex}
	\subfigure[]{
		\label{fig:eva1:sub3}
		\includegraphics[width=0.23\textwidth]{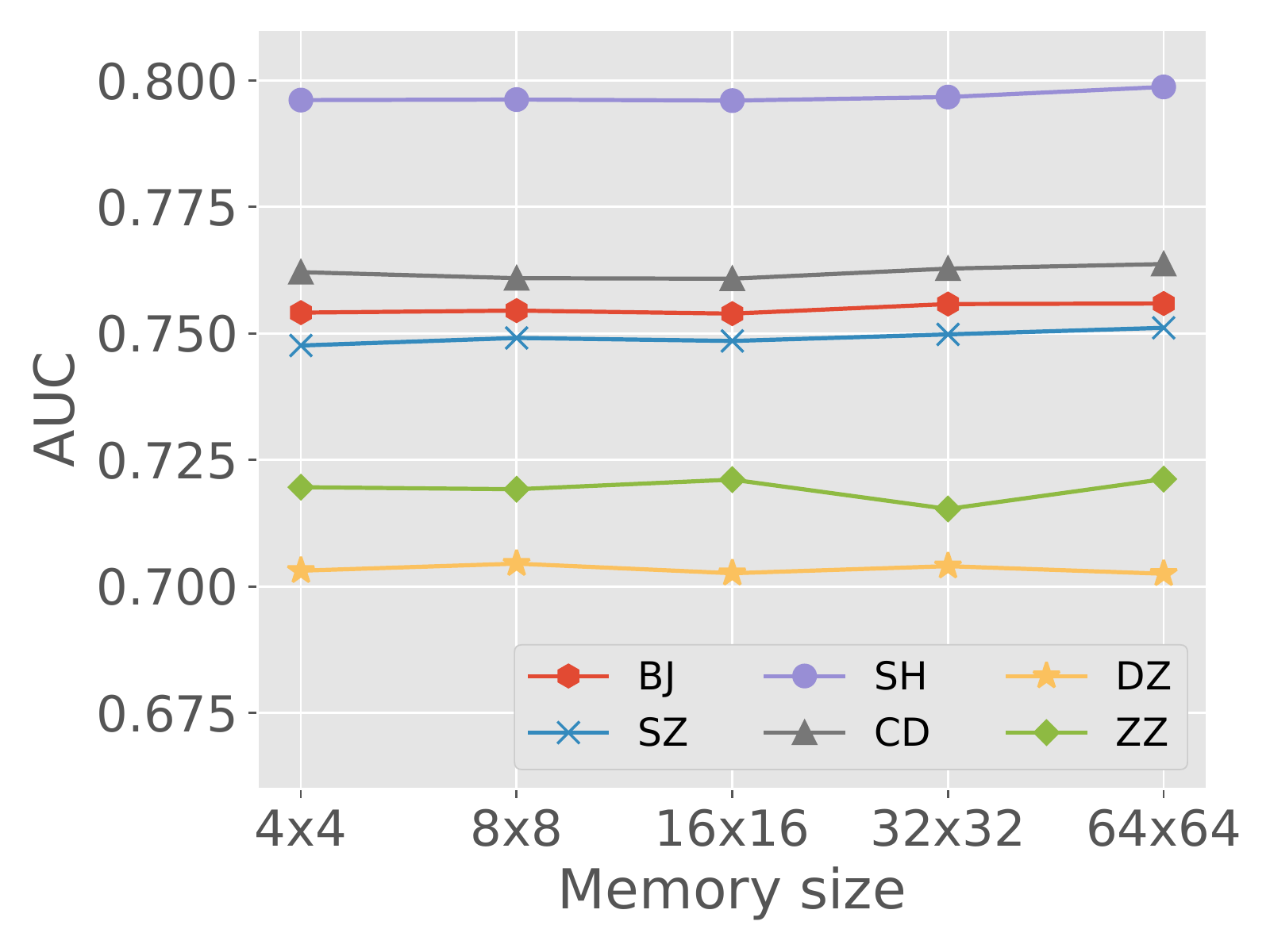}}
	\hspace{1ex}
	\subfigure[]{
		\label{fig:eva1:sub4}
		\includegraphics[width=0.23\textwidth]{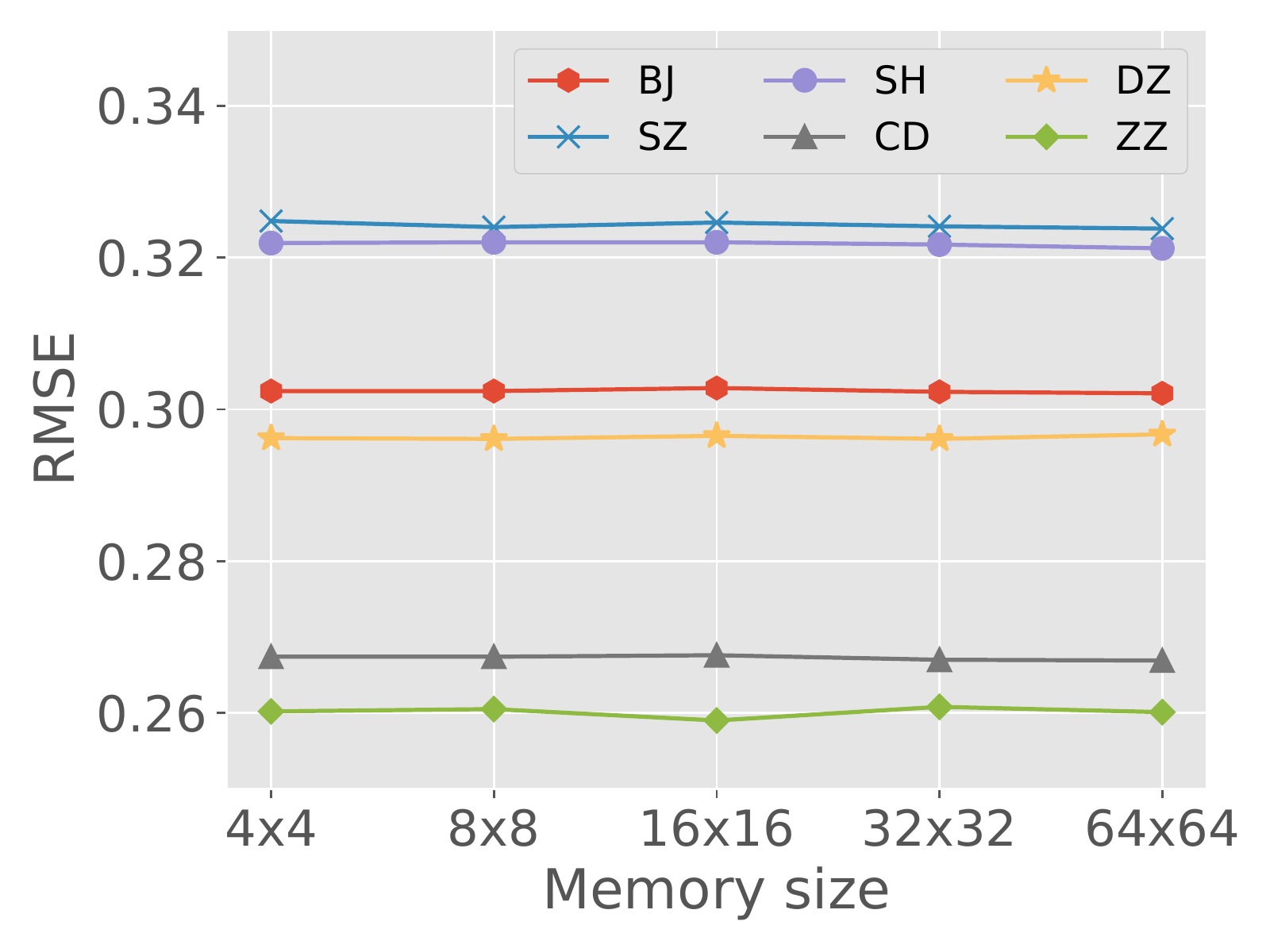}}
	%	\vspace{-2ex}
	\caption{Results of MV with different hyperparameter values.}
	\label{fig:eva1} %% label for entire figure
	%	\vspace{-4ex}
\end{figure*}

\subsection{Effectiveness of MV (RQ1)}
We first verify the performance of our MV model. Though we lay more emphasis on the accurate prediction for less popular cities (i.e., target cities) with sparser data, we test perform performance evaluation on all cities in this test. The upper part of Table \ref{table:overall-mv} shows the results of effectiveness comparison between MV and all feature interaction-based methods. Note that we have fully tuned all hyperparameters of all baselines with the Grid Search method. Hence, the results are recorded from their best performance. In this regard, we can ensure that the performance gain is from the model design itself. Also, our MSR prediction task shares a similar goal with user response prediction, where an improvement of our evaluation metrics at the 0.001-level is regarded significant~\cite{han2019all, song2019autoint} owing to the amount of additional revenue it can bring.

Based on the results, we draw the following observations from the experimental results:
\begin{itemize}
	\item It is obvious that MV outperforms all baselines on both AUC and RMSE significantly and consistently. Among the baselines, PNN has better performance than DeepFM which trivially fuse the output from the FM and DNN components. Other feature interaction models with more delicate structures (e.g., DIN which employs GRU to capture features' dynamics, DSSM which learns feature representations from different views, etc.) perform better than DeepFM and PNN. In addition, while Wide\&Deep and DeepCrossing have relative simple structures, they both show promising performance at the same time, which implies their generalizability.
	\item Generally, all methods show higher prediction accuracy. Moreover, the performance of DeepCrossing is poor on both target cities than other method while it outperforms all other baselines on source cities, which may indicate this method is not suitable for sparse datasets. Meanwhile, Wide\&Deep shows impressive prediction accuracy on target cities.
\end{itemize}

\subsection{Ablation Study on MV (RQ2)} 
In order to testify the importance of the key components of MV, we design several variants of it and conduct ablation study. The lower part of Table \ref{table:overall-mv} shows the results of the following 4 MV variants: (1) \textbf{MV-S1} that removes the interactions between features from different views in Section \ref{sec:contextbuilding}; (2) \textbf{MV-S2} that replaces the attentive aggregation in Section \ref{sec:interaction} by a simple concatenation operation; (3) \textbf{MV-S3} that replaces the DNC by a simple GRU network without the external memory; and (4) \textbf{MV-S4} that uses LSTM to replace the GRU in the DNC component. With the results obtained in Table \ref{table:overall-mv}, it can be observed that:
\begin{itemize}
	\item MV-S3 shows the worst performance overall, and it is also very unstable. The general performance of MV-S2 is the second worst, while MV-S1 and MV-S4 perform slightly better than MV-S2. Correspondingly, the DNC component plays the most important role to improve our model’s performance. Meanwhile, the attentive aggregation and the multi-view feature interaction both provide positive contributions to the prediction accuracy of the full MV model. 
	\item As can be expected, MV-S4 yields similar performance with MV on all source cities, as it shares the same components with MV except that the GRU in the DNC component is swapped with LSTM. It worth mentioning that MV-S4 performs not as well on the target cities as MV. One potential reason may be that the memory units within LSTM has more parameters, and the sparse data of target cities is insufficient to support its optimization.	
\end{itemize}

In summary, all the key components are beneficial for our model's performance, while the DNC makes the most contribution owing to its external memory. 
%	Through the above analyses, it is clear that most of the performance gain of MV comes from the DNC component in Section \ref{sec:contextbuilding}, which also proves that the context of the passenger-driver pair at that moment affects their decisions a lot. Then the attentive aggregation in Section \ref{sec:interaction} is the second contributor, which means finding the right factor combination is very important. 
%	These two points are our main advantages compared with baselines.

\subsection{Hyperparameter Sensitivity of MV (RQ3)}\label{sec:5.4.3}
As Fig. \ref{fig:eva1} shows, we examine the impact of two key hyperparameters of MV, namely the embedding dimension and the size of the memory matrix. In general, our model is insensitive to the variations of both hyperparameters, especially on larger datasets (i.e., source cities). We analyze the effect of each hyperparameter below:
\begin{itemize}
	\item{\textbf{Embedding Size:}} As Fig. \ref{fig:eva1:sub1} and Fig. \ref{fig:eva1:sub2} show, when the embedding size of MV is increasing, its performance keeps improving slightly on source cities, while a downward trend is observed on target cities. The possible reason might be that, source cities has richer data samples, hence a larger memory capacity is required to store all the information. Conversely, if the embedding size is set too large on target cities that have scarce information, it can lead to overfitting which would disturb the accuracy of the model. 	
	\item{\textbf{Memory Size:}} From Fig. \ref{fig:eva1:sub3} and Fig. \ref{fig:eva1:sub4}, we can see that the performance of MV on source cities also keeps increasing. It demonstrates the similar assumption in our experiments w.r.t. the embedding size on source cities. Compared with the embedding size, it seems that the memory size has little influence on the final prediction results. As suggested by the results, a relatively small memory matrix is sufficient for storing all the knowledge acquired from the dataset, which can also help avoid an excessively large model. With the increase of the memory size, the final results are relatively stable.
	%	Compared with different embedding sizes, the experimental results w.r.t. different memory sizes on target cities have slightly more fluctuations. It might be because that, the memory size is growing quadratically, causing more instability of the model performance.
\end{itemize}

\begin{table}[t]
	\renewcommand\arraystretch{1.3}
	\setlength{\abovecaptionskip}{0.1cm}
	\setlength\tabcolsep{8pt}
	\centering
	\caption{Results of different methods on target cities compared with KD.}
	\vspace{1ex}
	\begin{tabular}{|c|c|c|c|c|}
		\hline
		\multirow{2}*{Method}&\multicolumn{2}{c|}{DZ}&\multicolumn{2}{c|}{ZZ}\\ \cline{2-5}
		&AUC&RMSE&AUC&RMSE\\ \hline
		Single-FT (BJ)& 0.7131	&0.2945	&0.7292	&0.2568    \\
		Single-FT (SZ)& 0.7107	&0.2948	&0.729	&0.2571     \\
		Single-FT (SH)&0.7107	&0.2946	&0.7281	&0.2569    \\
		Single-FT (CD)&0.7088	&0.2944	&0.7282	&0.2569   \\
		Muti-FT & 0.7122	&0.294	 &0.7303	 &0.2566   \\
		\textbf{MV$^+$}& 0.7131 &0.2947 &0.7402	&0.2532    \\ 
		\textbf{MV$^-$}&\textbf{0.7136}  & \textbf{0.2935}     & \textbf{0.7425} &\textbf{ 0.2526 }  \\ \hline
	\end{tabular}
	\label{table:overall-kd}
\end{table}

\subsection{Effectiveness of KD (RQ4)}
The KD model is designed to cope with the significantly scarcer data in target cities using the knowledge from source cities. Table \ref{table:overall-kd} compares the performance of all knowledge transfer-based baseline methods and KD. In Table \ref{table:overall-kd}, we have listed the performance of both the complex teacher model (i.e., MV$^+$) and the simplified student model (i.e., MV$^-$) trained via KD. We conclude our findings below: 
\begin{itemize}
	\item Combing the results from Table \ref{table:overall-kd} with Table \ref{table:overall-mv}, we can tell that transferring knowledge from source cities to target cities can make an obvious difference in performance compared with merely modelling feature interactions. The superiority of MV$^-$ verifies the effectiveness of the KD framework that we have proposed, where it even slightly outperforms the sophisticated teacher model on both target cities.	
	\item As Table\ref{table:overall-kd} shows, Single-FT based on BJ dataset is better than the other three Single-FT baselines. One possible reason is that BJ owns the most sufficient passenger-driver matching records, which will intuitively offer richer knowledge to the target cites during inference. At the same time, Multi-FT performs better than Single-FT in general, indicating the knowledge from a single source city is hardly sufficient for predicting MSR on target cities. 
\end{itemize}

\begin{figure*}[t!]
	\centering
	\setlength{\abovecaptionskip}{0.10cm}
	\subfigure[]{
		\label{fig:eva2:sub1}
		\includegraphics[width=0.23\textwidth]{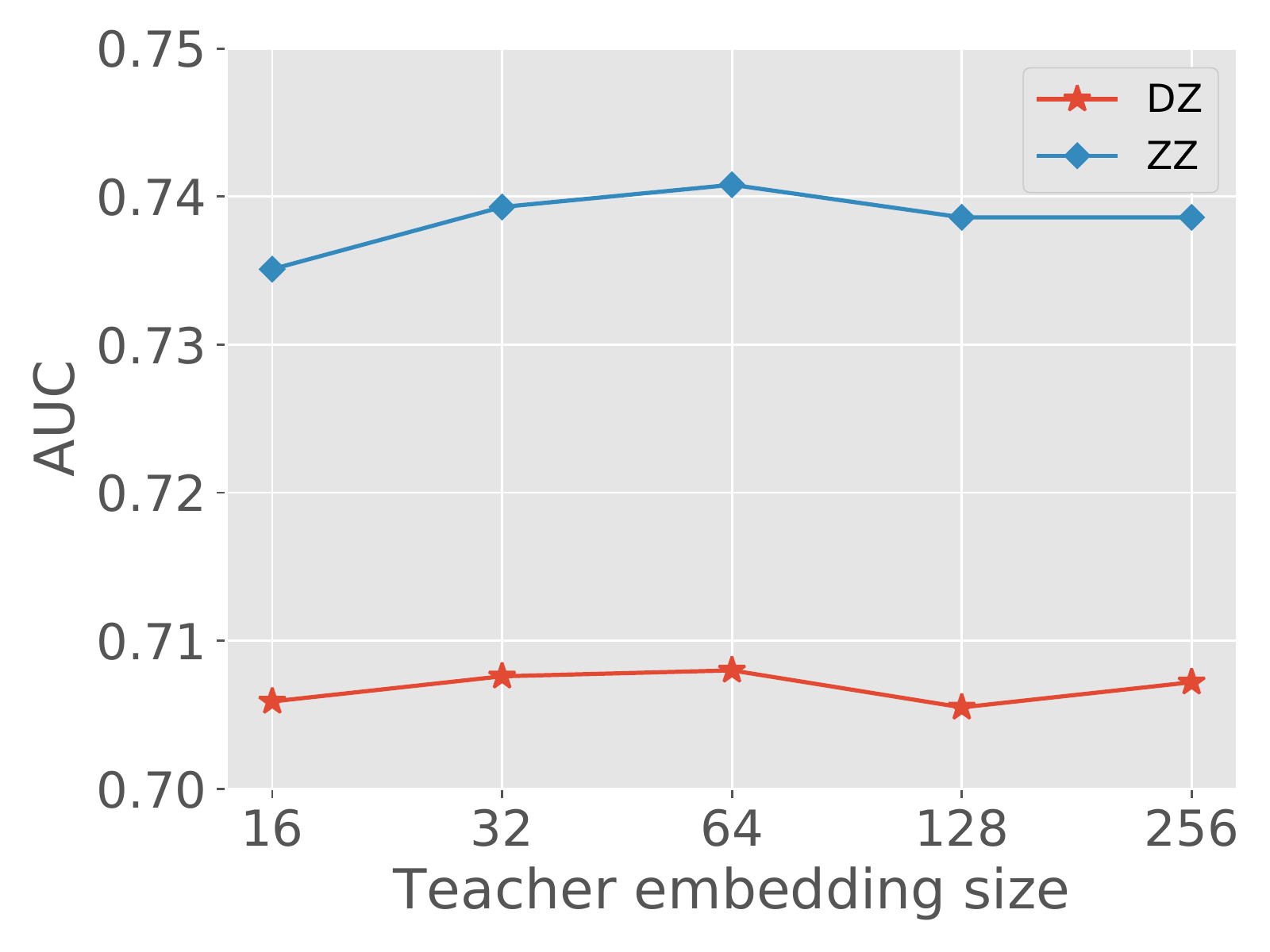}}
	\hspace{1ex}
	\subfigure[]{
		\label{fig:eva2:sub2}
		\includegraphics[width=0.23\textwidth]{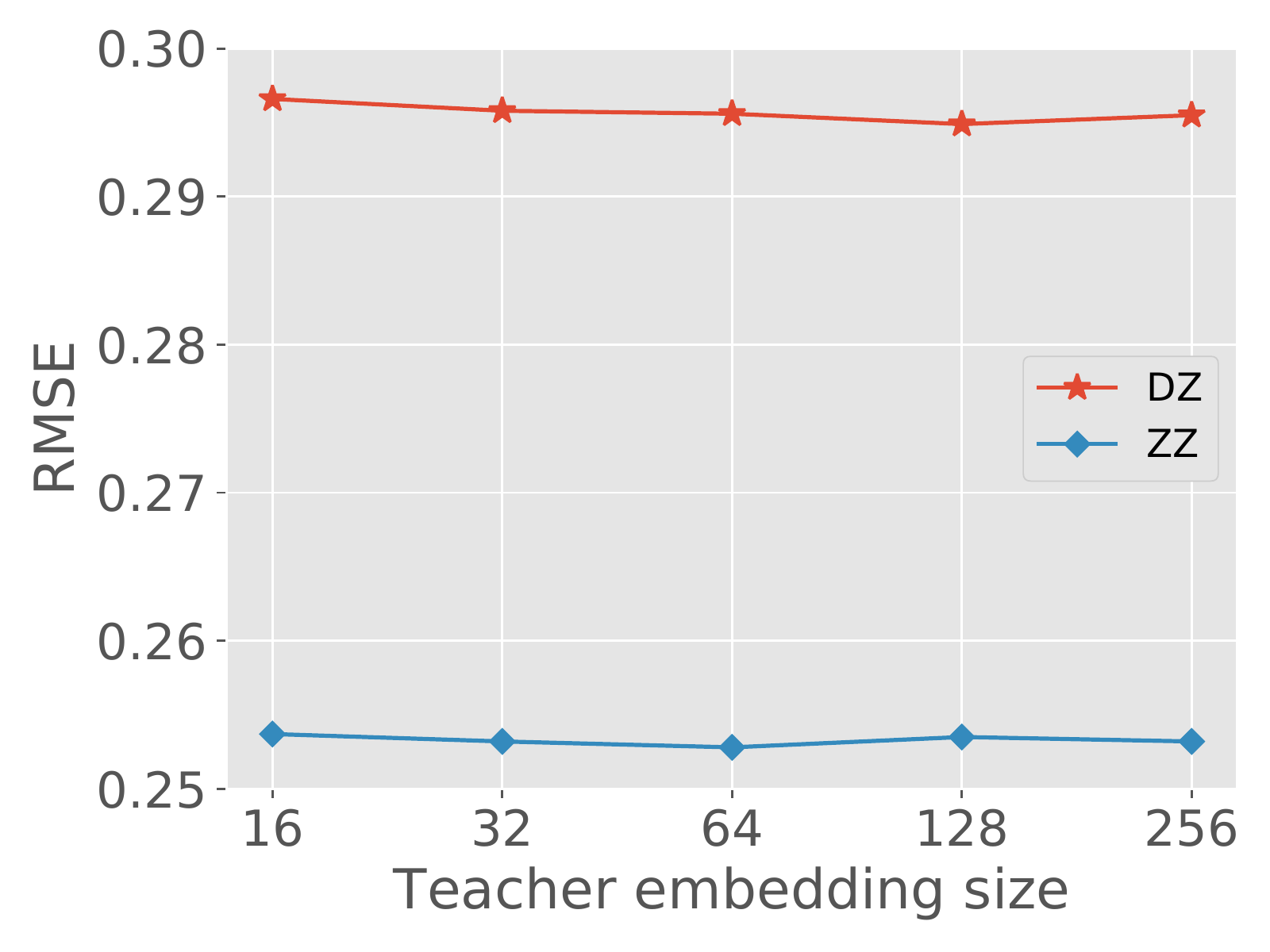}}
	\hspace{1ex}
	\subfigure[]{
		\label{fig:eva2:sub3}
		\includegraphics[width=0.23\textwidth]{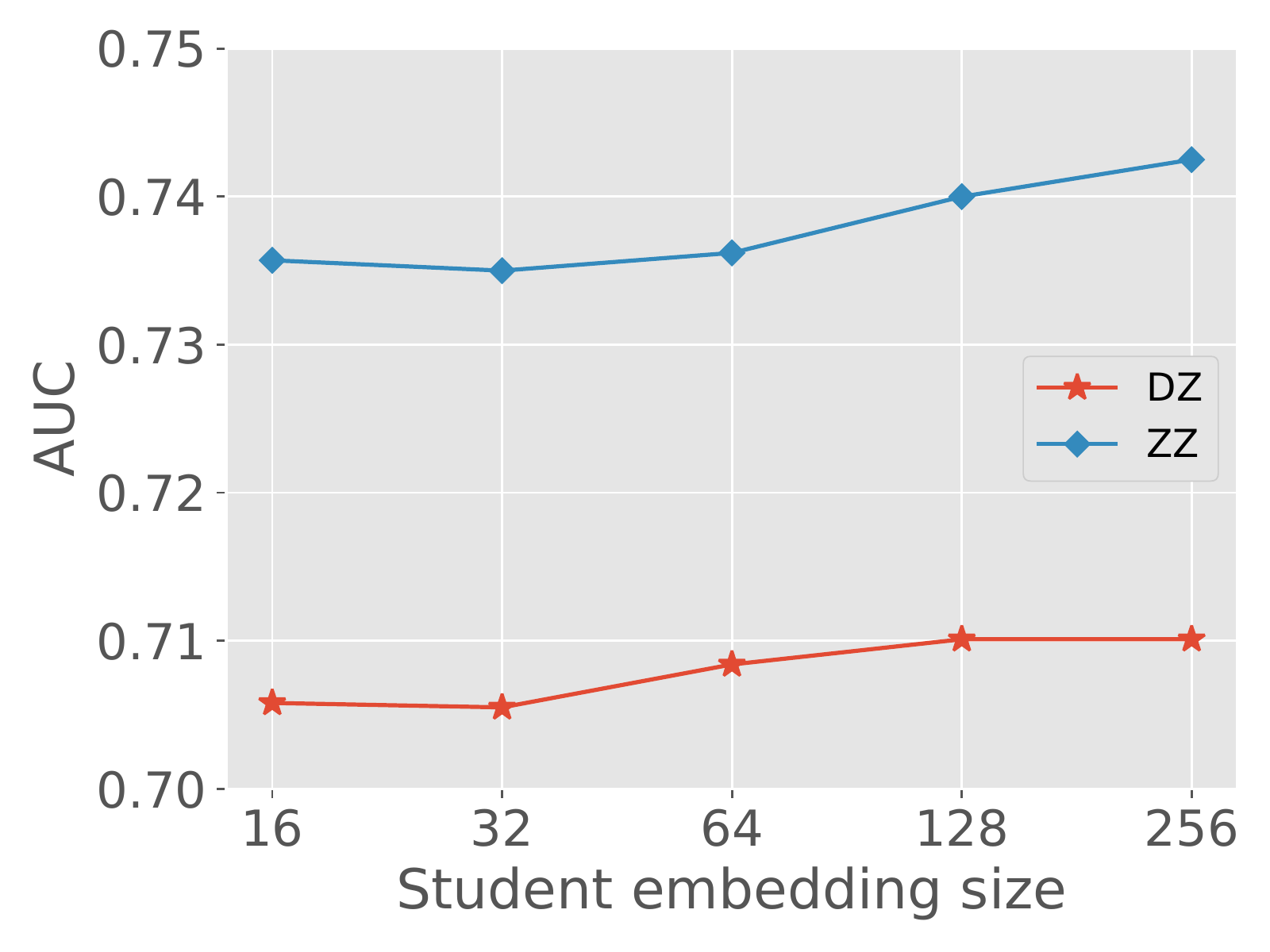}}
	\hspace{1ex}
	\subfigure[]{
		\label{fig:eva2:sub4}
		\includegraphics[width=0.23\textwidth]{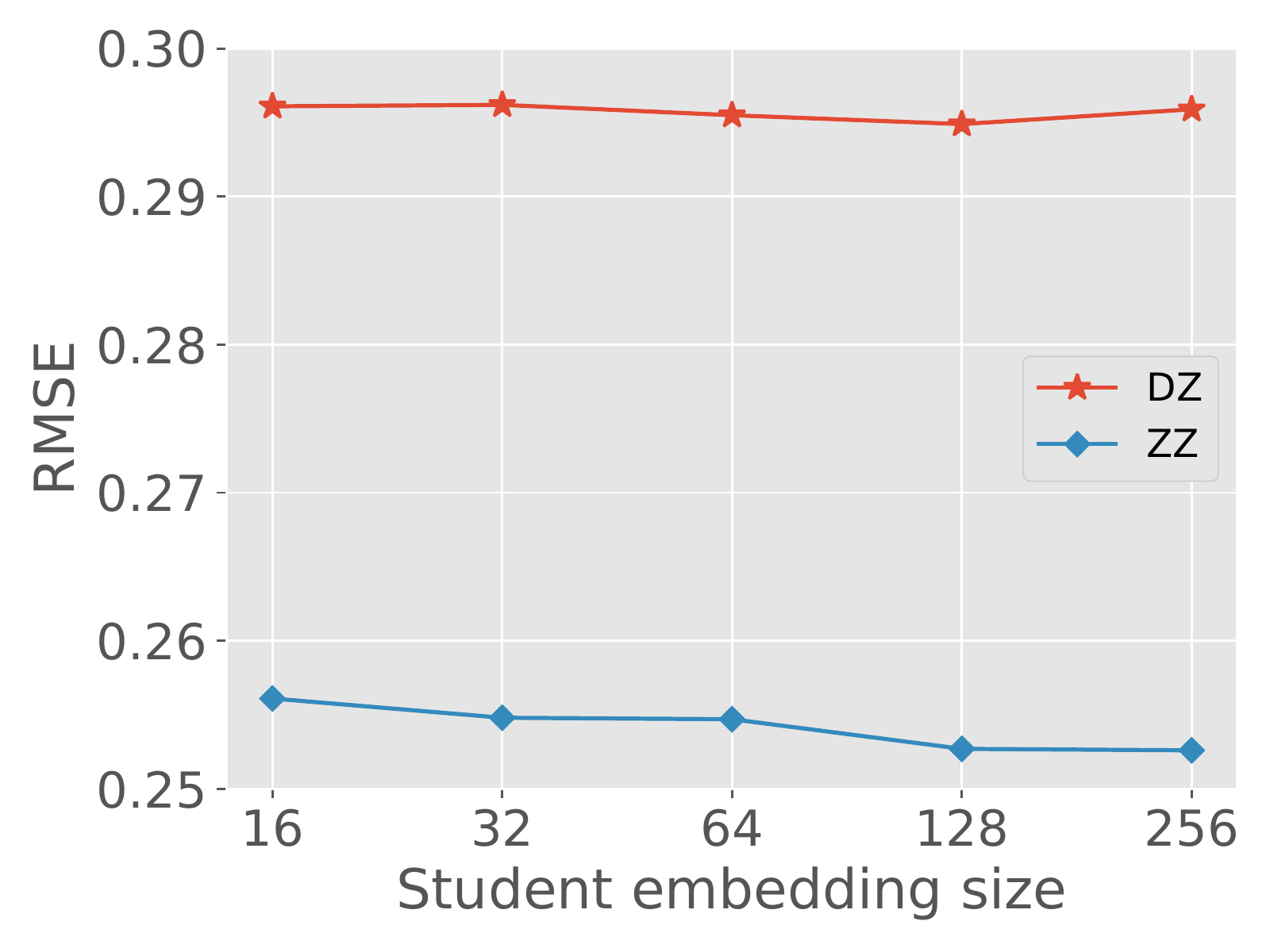}}
	%	\vspace{-2ex}
	\caption{Results of KD with different hyperparameter values.}
	\label{fig:eva2} %% label for entire figure
	%	\vspace{-4ex}
\end{figure*}

\subsection{Hyperparameter Sensitivity of KD (RQ5)}
As KD takes the learned memory matrix of each city's MV model as its input, the memory size of KD is already determined. Hence, we focus on how different the embedding sizes in KD influence the final prediction performance. Fig. \ref{fig:eva2} depicts the prediction results of KD with different values of embedding sizes, and we perform experiments for both the teacher and student component. Compared with the MV in Section \ref{sec:5.4.3}, KD is more sensitive to the embedding dimension. We hereby provide detailed analysis on the hyperparameter sensitivity of KD:
\begin{itemize}
	\item{\textbf{Embedding Size of Teacher Model:}} As Fig. \ref{fig:eva2:sub1} and Fig. \ref{fig:eva2:sub2} show, when the embedding size grows, the performance of KD rises at first and then gradually declines. KD performs the best when the teacher embedding size is set to $64$. One possible reason might be that, the teacher model is fully supported by the comprehensive knowledge from all source cities, so a moderate value of $64$ is sufficient for the embedding size. Another potential reason is that, similar to the MV in Section \ref{sec:5.4.3}, the scarce data of target cities means that the model does not need a very large embedding size to learn all the information .	
	\item{\textbf{Embedding Size of Student Model:}} Fig. \ref{fig:eva2:sub3} and Fig. \ref{fig:eva2:sub4} show that the KD's performance is constantly improving with the increase of student embedding size. The rationale of why it shows a different trend compared with the teacher embedding size is that, the student model is considerably simpler and has much less parameters. Therefore, it benefits from a larger embedding size to facilitate expressiveness so that the student can fully replicate the behavior of the accurate teacher model.
\end{itemize}

\begin{figure*}[tb!]
	\centering
	\setlength{\abovecaptionskip}{0.10cm}
	\subfigure[]{
		\label{fig:eva3:sub1}
		\includegraphics[width=0.3\textwidth]{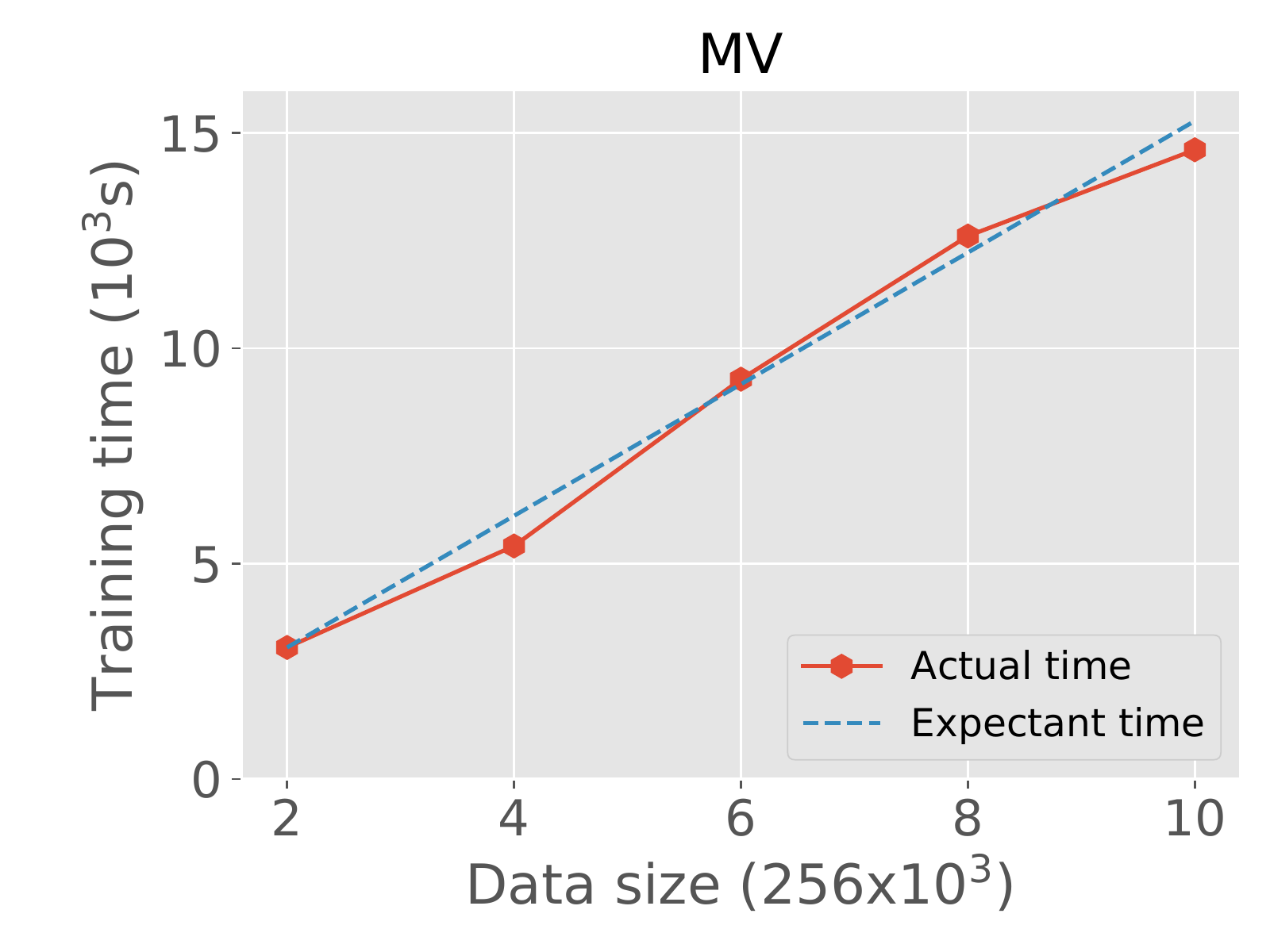}}
	\hspace{1ex}
	\subfigure[]{
		\label{fig:eva3:sub2}
		\includegraphics[width=0.3\textwidth]{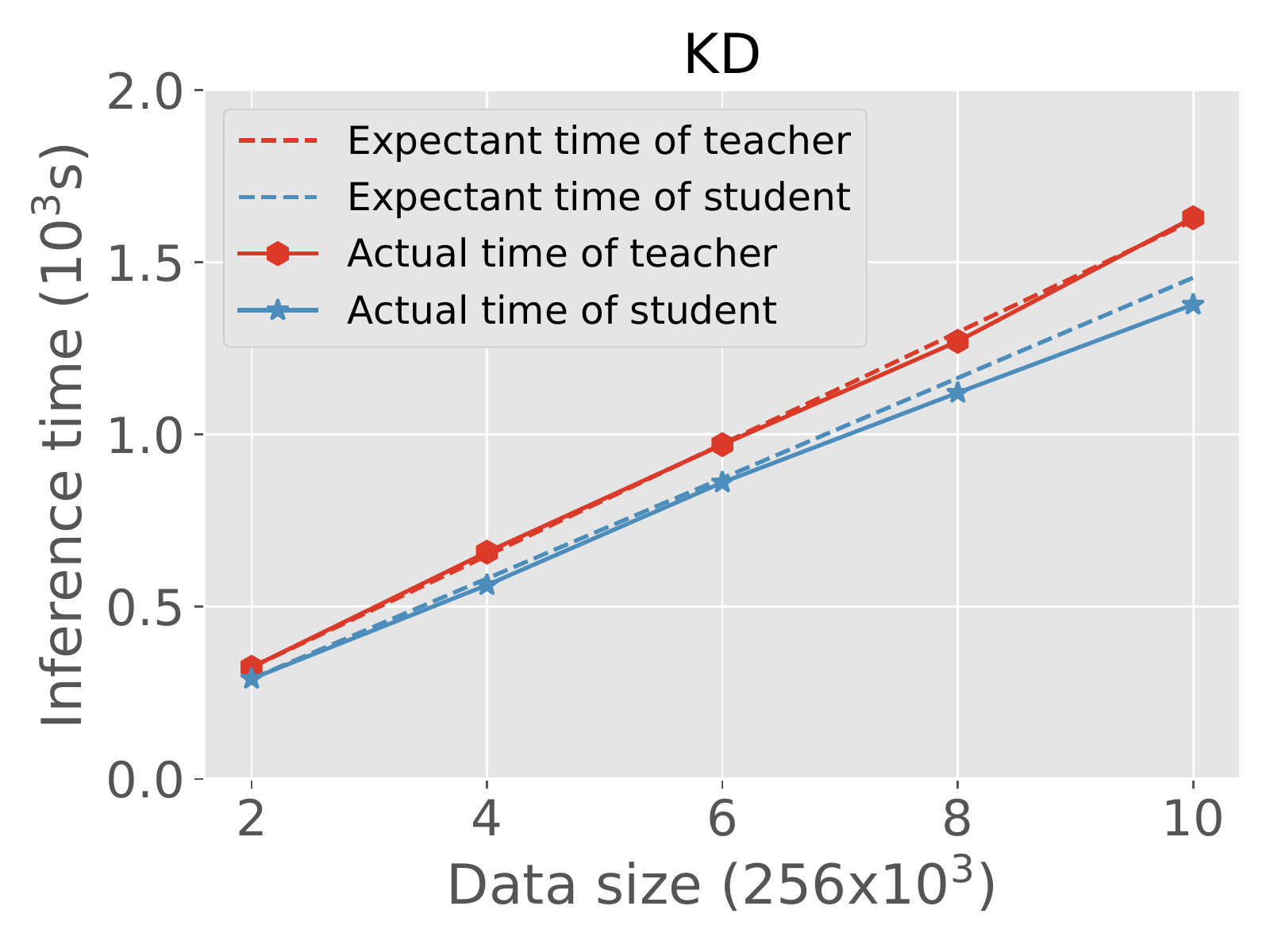}}
	\hspace{1ex}
	\subfigure[]{
		\label{fig:eva3:sub3}
		\includegraphics[width=0.3\textwidth]{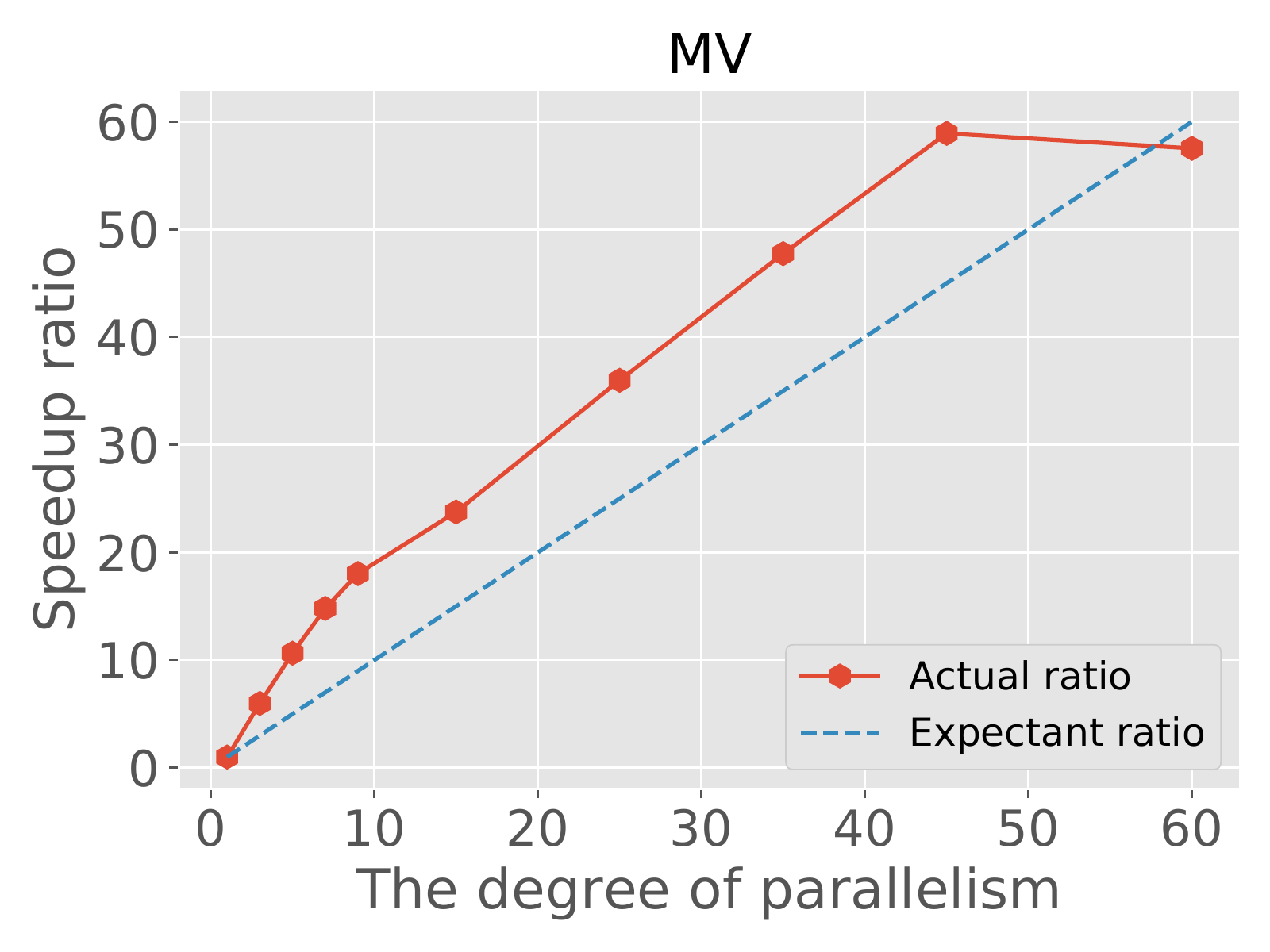}}
	\caption{Scalability analysis results.}
	\label{fig:eva3} %% label for entire figure
\end{figure*}

\subsection{Scalability Analysis of MV and KD (RQ6)} 
We test the scalability of MV and KD via three groups of experiments and the results is shown in Fig. \ref{fig:eva3}. In what follows, we present more details on the experimental settings and provide further analysis on our model's scalability.
\begin{itemize}
	\item Firstly, we test the training efficiency of MV by varying the data size used for training. Specifically, we train MV with different amounts of instances, and report the time it takes for model convergence. We can see from Fig. \ref{fig:eva3:sub1} that the training time for MV oscillates around the linear one (blue line) very closely. This indicates that MV is able to scale to large-scale datasets.
	\item Secondly, recall that the other important goal of designing KD is to obtain an efficient and simple model to support the online inference while preserving high accuracy. Hence, we compare the time cost of teacher and student models in KD during inference. As the curves in Fig. \ref{fig:eva3:sub2} show, the inference efficiency of student is constantly better than the teacher, and both teacher and student models' inference time costs grow linearly.
	\item Thirdly, as the MSR prediction model is expected to handle very large datasets during production, we deploy our model on a spark cluster with parallelism for training. The experiments of this group are conducted under the same data size of $256 \times 10^4$ instances. The speed-up ratio of MV under different degrees of parallelism is depicted in Fig. \ref{fig:eva3:sub3}. It is obvious that the efficiency of parallelism is highly competitive. Moreover, we can see that the speed-up ratio begins to drop after the degree of parallelism increases to $45$, which is a strong efficiency indicator for industry-level deployment.
\end{itemize}

\section{RELATED WORK}\label{sec:related}
\subsection{Traffic Prediction}
We first introduce relevant research work on traffic prediction problems~\cite{wei2016zest, wang2016traffic, tong2017simpler, liu2017functional, zhang2017deep, feng2018deepmove, yao2018deep, yao2019learning, wang2019unified, wang2019origin}. With the fast development of Internet and data warehousing techniques, all kinds of data is becoming increasingly available for such prediction tasks in the last few years. Some studies~\cite{wei2016zest, wang2019unified, tong2017simpler, yao2018deep} propose to use multi-source data to solve the passenger demand prediction. Wang and Wei et al.~\cite{wei2016zest, wang2019unified} present a combinatorial model to predict the passenger demand in a region at the future time slot, and the model captures the temporal trends and fuses the spatial and other related features to facilitate prediction.
%Yao et al.~\cite{yao2018deep} propose a deep multi-view spatial-temporal network to model both spatial and temporal relations. Specifically, their model consists of three views: temporal view (modeling correlations between future demand values with near time points via LSTM), spatial view (modeling local spatial correlation via local CNN), and semantic view (modeling correlations among regions sharing similar temporal patterns).
In particular, Tong et al.~\cite{tong2017simpler} employ multi-source data to synthesize more than 200 million features via feature engineering and feed them into a unified linear regression model to obtain prediction results. Zhang et al.~\cite{zhang2017deep} design a spatiotemporal model based on CNN and residual network to extract information from both spatial and temporal perspectives and predict both the in-flow and out-flow within each urban area. 
Wang et al.~\cite{wang2016traffic} propose a deep learning method with an error-feedback recurrent convolutional neural network (eRCNN) for continuous traffic speed prediction.
Liu et al.~\cite{liu2017functional} develop a hierarchical station bike demand predictor which analyzes bike demands from functional zone level to station level. However, these studies ignore the data imbalance problem among different cities. Yao et al.~\cite{yao2019learning} propose a meta-learning framework to transfer knowledge from source cities to target cities, but their model does not account for intricate feature interactions and the efficiency of the prediction model. 
Moreover, Tong et al.'s work provide us with much inspiration, especially studies on bipartite graph matching \cite{DBLP:conf/icde/WangTLXXL19} and task assignment problem \cite{DBLP:conf/kdd/ShiTZSLY21, tong2019two, tong2016mobile} and online minimum matching problem in real-time spatial data~\cite{tong2016online}. Though these are also two-sided problems, such methods lack the capability of addressing issues of data imbalance and efficient online inference in the MSR prediction context.

\subsection{Deep Feature Interaction Models}
As the MSR problem involves the joint effect of multiple dynamic factors, modeling feature interactions is necessary. Recently popular methods on feature interaction modeling~\cite{shan2016deep, qu2018product, zhou2018deep, huang2013learning, shen2014latent, palangi2014semantic, elkahky2015a, Rendle2010factorization, juan2016field, guo2017deepfm, he2017neural, xiao2017attentional, lian2018xdeepfm, chen2020sequence} have gained immense popularity in a wide range of prediction tasks, especially user response prediction in online advertising. 
In order to project queries and documents into a common low-dimensional space where the relevance of a document given a query is readily
computed as the distance between them, Huang et al.~\cite{huang2013learning} propose a DSSM which learns representations for both objects and computes their relevance. Similarly structured methods~\cite{shen2014latent, palangi2014semantic, elkahky2015a} based on CNN, LSTM are proposed aiming at capturing more interaction contexts. 
Some studies~\cite{juan2016field, guo2017deepfm, he2017neural, xiao2017attentional, lian2018xdeepfm, chen2020sequence} focus on improving the interaction modeling schemes based on the original factorization machine (FM)~\cite{Rendle2010factorization}. 
For example, Guo et al.~\cite{guo2017deepfm} combines FM and the traditional DNN, which enhances the expressiveness of the model. Xiao et al.~\cite{xiao2017attentional} propose a novel attentional factorization machine, which learns the importance of each feature interaction via a neural attention network. 
Qu et al.~\cite{qu2016product, qu2018product} come up with a product-based neural network (PNN), which performs either inner product or outer product operation on feature embeddings.
Zhou et al.~\cite{zhou2018deep} calculate the attention on different features and result in a weighted sum in different feature groups, which enables the model to adjust the weights of input features for accurate prediction.
Chen et al.~\cite{chen2020sequence} point out that existing FM-based models assume no temporal orders in the data, and are unable to capture the sequential dependencies or patterns within the dynamic features, so they propose a novel sequence-aware factorization machine for temporal predictive analytics, which models feature interactions by fully investigating the effect of sequential dependencies. Unfortunately, all aforementioned methods do not take the data imbalance problem into consideration, which greatly hinders a model's performance under data scarcity.

\subsection{Memory Networks and Knowledge Distillation}
In order to solve the data imbalance problem, we gain lots of inspiration from memory networks and knowledge distillation methods~\cite{graves2014neural, graves2016hybrid, cheng2016wide, ba2013deep, hinton2015distilling, mishra2017apprentice, tang2018ranking, wang2020next}. One representative technique of  is memory networks whose theory is learning from previous experience. Graves et al.~\cite{graves2014neural, graves2016hybrid} propose NTM and DNC which maintain an external memory matrix to store the dynamics of temporal features over time. Then, in order to balance the generalization and memorization of the model, Cheng et al.~\cite{cheng2016wide} design the Wide\&Deep model, whose wide part is in charge of the memorization ability via a linear model and the deep part is responsible for the generalization of the model. As a popular choice for transfer learning, knowledge distillation has shown promising effectiveness~\cite{ba2013deep, hinton2015distilling, mishra2017apprentice, tang2018ranking, wang2020next} in various tasks. Ba et al.~\cite{ba2013deep} is the first to put forward the emulation learning scheme where a complicated teacher model is designed to guide the training of a simple student model without labels, which has been attracting extensive research attention until now. Hinton et al.~\cite{hinton2015distilling} propose to further incorporate the labels of the training data into the loss function to enhance the performance of the learned student model. Knowledge distillation has been proven effective in recent preference mining tasks like recommendation~\cite{tang2018ranking, wang2020next}, showcasing its potential in MSR prediction.

\section{CONCLUSION}\label{sec:conclusion}
In this work, we define a novel research problem, i.e.,  \underline{\textbf{M}}atching \underline{\textbf{S}}uccess \underline{\textbf{R}}ate prediction for passenger-driver pairs, which originates from the real demand of ride-hailing platforms. MSR prediction face three main challenges. Firstly, as MSR involves a bidirectional decision process of two end-users in a dynamic environment, learning a comprehensive representation for each passenger-driver pair is non-trivial. Secondly, the data imbalance problem is common yet harmful for the prediction accuracy on small cities. Thirdly, as MSR prediction is utilized to support real-time strategic operations, a lightweight yet accurate model is essential. In order to solve MSR prediction, we propose the \underline{\textbf{M}}ulti-\underline{\textbf{V}}iew model (\textbf{MV}) which learns the combinatorial effect of features in different views. Then, to tackle data imbalance and guarantee online efficiency, we design the \underline{\textbf{K}}nowledge \underline{\textbf{D}}istillation framework (\textbf{KD}) which can not only supplement knowledge for the cities with scarce data, but also generate a simple model to support online applications. Through extensive experiments, we have demonstrated the strength of our solution in both accuracy and scalability.

\section*{Acknowledgement}
Thanks to my dear colleagues in Didi Chuxing, Kecheng Xu and Haoyu Wang who help us conduct data statistics during the first-round revision. This work is supported by the National Key Research \& Development Program of China (Grant No. 2016YFB1000103) and Australian Research Council (Grant No. DP190101985, FT210100624).

\bibliography{MSR}

\begin{thebibliography}{10}
\providecommand{\url}[1]{#1}
\csname url@samestyle\endcsname
\providecommand{\newblock}{\relax}
\providecommand{\bibinfo}[2]{#2}
\providecommand{\BIBentrySTDinterwordspacing}{\spaceskip=0pt\relax}
\providecommand{\BIBentryALTinterwordstretchfactor}{4}
\providecommand{\BIBentryALTinterwordspacing}{\spaceskip=\fontdimen2\font plus
\BIBentryALTinterwordstretchfactor\fontdimen3\font minus
  \fontdimen4\font\relax}
\providecommand{\BIBforeignlanguage}[2]{{%
\expandafter\ifx\csname l@#1\endcsname\relax
\typeout{** WARNING: IEEEtran.bst: No hyphenation pattern has been}%
\typeout{** loaded for the language `#1'. Using the pattern for}%
\typeout{** the default language instead.}%
\else
\language=\csname l@#1\endcsname
\fi
#2}}
\providecommand{\BIBdecl}{\relax}
\BIBdecl

\bibitem{qu2016product}
Y.~Qu, H.~Cai, K.~Ren, W.~Zhang, Y.~Yu, Y.~Wen, and J.~Wang, ``Product-based
  neural networks for user response prediction,'' \emph{2016 IEEE 16th
  International Conference on Data Mining (ICDM)}, pp. 1149--1154, 2016.

\bibitem{huang2013learning}
P.-S. Huang, X.~He, J.~Gao, L.~Deng, A.~Acero, and L.~Heck, ``Learning deep
  structured semantic models for web search using clickthrough data,'' in
  \emph{Proceedings of the 22nd ACM International Conference on Information and
  Knowledge Management}, 2013, p. 2333–2338.

\bibitem{liu2017functional}
J.~Liu, L.~Sun, Q.~Li, J.~Ming, Y.~Liu, and H.~Xiong, ``Functional zone based
  hierarchical demand prediction for bike system expansion,'' in
  \emph{Proceedings of the 23rd ACM SIGKDD International Conference on
  Knowledge Discovery and Data Mining}, 2017, pp. 957--966.

\bibitem{zhang2017deep}
J.~Zhang, Y.~Zheng, and D.~Qi, ``Deep spatio-temporal residual networks for
  citywide crowd flows prediction,'' in \emph{Proceedings of the AAAI
  Conference on Artificial Intelligence}, vol.~31, no.~1, 2017.

\bibitem{wang2019unified}
Y.~Wang, X.~Lin, H.~Wei, T.~Wo, Z.~Huang, Y.~Zhang, and J.~Xu, ``A unified
  framework with multi-source data for predicting passenger demands of ride
  services,'' \emph{ACM Transactions on Knowledge Discovery from Data (TKDD)},
  vol.~13, no.~6, pp. 1--24, 2019.

\bibitem{yao2019learning}
H.~Yao, Y.~Liu, Y.~Wei, X.~Tang, and Z.~Li, ``Learning from multiple cities: A
  meta-learning approach for spatial-temporal prediction,'' in \emph{The World
  Wide Web Conference}, 2019, pp. 2181--2191.

\bibitem{wang2018crowd}
L.~Wang, X.~Geng, X.~Ma, F.~Liu, and Q.~Yang, ``Crowd flow prediction by deep
  spatio-temporal transfer learning,'' \emph{arXiv preprint arXiv:1802.00386},
  2018.

\bibitem{luo2020dynamic}
W.~Luo, H.~Zhang, X.~Yang, X.~Y. Lin~Bo, Z.~Li, X.~Qie, and J.~Ye, ``Dynamic
  heterogeneous graph neural network for real-time event prediction,'' in
  \emph{Proceedings of the 26rd ACM SIGKDD international conference on
  knowledge discovery and data mining}, 2020, pp. 3213--3223.

\bibitem{guo2017deepfm}
H.~Guo, R.~Tang, Y.~Ye, Z.~Li, and X.~He, ``Deepfm: A factorization-machine
  based neural network for ctr prediction,'' in \emph{Proceedings of the 26th
  International Joint Conference on Artificial Intelligence (IJCAI)}, 2017, p.
  1725–1731.

\bibitem{juan2016field}
Y.~Juan, Y.~Zhuang, W.-S. Chin, and C.-J. Lin, ``Field-aware factorization
  machines for ctr prediction,'' in \emph{Proceedings of the 10th ACM
  Conference on Recommender Systems}, 2016, p. 43–50.

\bibitem{shen2014latent}
Y.~Shen, X.~He, J.~Gao, L.~Deng, and G.~Mesnil, ``A latent semantic model with
  convolutional-pooling structure for information retrieval,'' in
  \emph{Proceedings of the 23rd ACM international conference on conference on
  information and knowledge management}, 2014, pp. 101--110.

\bibitem{elkahky2015a}
A.~M. Elkahky, Y.~Song, and X.~He, ``A multi-view deep learning approach for
  cross domain user modeling in recommendation systems,'' in \emph{Proceedings
  of the 24th International Conference on World Wide Web}, 2015, p. 278–288.

\bibitem{he2017neural}
X.~He and T.-S. Chua, ``Neural factorization machines for sparse predictive
  analytics,'' in \emph{Proceedings of the 40th International ACM SIGIR
  Conference on Research and Development in Information Retrieval (SIGIR)},
  2017, p. 355–364.

\bibitem{lian2018xdeepfm}
J.~Lian, X.~Zhou, F.~Zhang, Z.~Chen, X.~Xie, and G.~Sun, ``Xdeepfm: Combining
  explicit and implicit feature interactions for recommender systems,'' in
  \emph{Proceedings of the 24th ACM SIGKDD International Conference on
  Knowledge Discovery and Data Mining}, 2018, p. 1754–1763.

\bibitem{palangi2014semantic}
H.~Palangi, L.~Deng, Y.~Shen, J.~Gao, X.~He, J.~Chen, X.~Song, and R.~Ward,
  ``Semantic modelling with long-short-term memory for information retrieval,''
  \emph{arXiv preprint arXiv:1412.6629}, 2014.

\bibitem{song2016multi}
Y.~Song, A.~M. Elkahky, and X.~He, ``Multi-rate deep learning for temporal
  recommendation,'' in \emph{Proceedings of the 39th International ACM SIGIR
  conference on Research and Development in Information Retrieval}, 2016, pp.
  909--912.

\bibitem{chen2020sequence}
T.~Chen, H.~Yin, Q.~V.~H. Nguyen, W.-C. Peng, X.~Li, and X.~Zhou,
  ``Sequence-aware factorization machines for temporal predictive analystics,''
  in \emph{IEEE 36th International Conference on Data Engineering (ICDE)},
  2020.

\bibitem{graves2016hybrid}
A.~Graves, G.~Wayne, M.~Reynolds1, T.~Harley1, I.~Danihelka1, and
  A.~Grabska-Barwińska1, ``Hybrid computing using a neural network with
  dynamic external memory,'' \emph{Nature}, vol. 538, pp. 471--491, 2016.

\bibitem{graves2014neural}
A.~Graves, G.~Wayne, and I.~Danihelka, ``Neural turing machine,'' \emph{arXiv
  preprint}, p. arXiv:1410.5401, 2014.

\bibitem{ba2013deep}
L.~J. Ba and R.~Caruana, ``Do deep nets really need to be deep?'' \emph{arXiv
  preprint arXiv:1312.6184}, 2013.

\bibitem{hinton2015distilling}
G.~Hinton, O.~Vinyals, and J.~Dean, ``Distilling the knowledge in a neural
  network,'' \emph{arXiv preprint arXiv:1503.02531}, 2015.

\bibitem{mishra2017apprentice}
A.~Mishra and D.~Marr, ``Apprentice: Using knowledge distillation techniques to
  improve low-precision network accuracy,'' \emph{arXiv preprint
  arXiv:1711.05852}, 2017.

\bibitem{hochreiter1997long}
S.~Hochreiter and J.~Schmidhuber, ``Long short-term memory,'' \emph{MIT Press
  Neural Computation.}, vol.~9, p. 1735–1780, 1997.

\bibitem{cho2014learning}
K.~Cho, B.~van Merrienboer, C.~Gulcehre, D.~Bahdanau, F.~Bougares, H.~Schwenk,
  and Y.~Bengio, ``Learning phrase representations using rnn encoder-decoder
  for statistical machine translation,'' \emph{arXiv preprint}, p.
  arXiv:1406.1078, 2014.

\bibitem{wang2017gated}
W.~Wang, N.~Yang, F.~Wei, B.~Chang, and M.~Zhou, ``Gated self-matching networks
  for reading comprehension and question answering.'' \emph{In Proceedings of
  the 55th Annual Meeting of the Association for Computational Linguistics
  (ACL)}.

\bibitem{kingma2014adam}
D.~P. Kingma and J.~Ba, ``Adam: A method for stochastic optimization,''
  \emph{arXiv preprint}, 2014.

\bibitem{zhou2018deep}
G.~Zhou, X.~Zhu, C.~Song, Y.~Fan, H.~Zhu, X.~Ma, Y.~Yan, J.~Jin, H.~Li, and
  K.~Gai, ``Deep interest network for click-through rate prediction,'' in
  \emph{Proceedings of the 24th ACM SIGKDD International Conference on
  Knowledge Discovery and Data Mining}, 2018, p. 1059–1068.

\bibitem{wang2017deep}
R.~Wang, B.~Fu, G.~Fu, and M.~Wang, ``Deep andcross network for ad click
  predictions,'' in \emph{Proceedings of the ADKDD'17}, 2017.

\bibitem{cheng2016wide}
H.-T. Cheng, L.~Koc, J.~Harmsen, T.~Shaked, T.~Chandra, H.~Aradhye,
  G.~Anderson, G.~Corrado, W.~Chai, M.~Ispir, R.~Anil, Z.~Haque, L.~Hong,
  V.~Jain, X.~Liu, and H.~Shah, ``Wide and deep learning for recommender
  systems,'' in \emph{Proceedings of the 1st Workshop on Deep Learning for
  Recommender Systems}, 2016, p. 7–10.

\bibitem{shan2016deep}
Y.~Shan, T.~R. Hoens, J.~Jiao, H.~Wang, D.~Yu, and J.~Mao, ``Deep crossing:
  Web-scale modeling without manually crafted combinatorial features,'' in
  \emph{Proceedings of the 22nd ACM SIGKDD International Conference on
  Knowledge Discovery and Data Mining}, 2016, p. 255–262.

\bibitem{han2019all}
Z.~Li, W.~Cheng, Y.~Chen, H.~Chen, and W.~Wang, ``Interpretable click-through
  rate prediction through hierarchical attention,'' in \emph{Proceedings of the
  Thirteenth ACM International Conference on Web Search and Data Mining}, 2020,
  p. 313–321.

\bibitem{song2019autoint}
W.~Song, C.~Shi, Z.~Xiao, Z.~Duan, Y.~Xu, M.~Zhang, and J.~Tang, ``Autoint:
  Automatic feature interaction learning via self-attentive neural networks,''
  in \emph{Proceedings of the 28th ACM International Conference on Information
  and Knowledge Management}, 2019, pp. 1161--1170.

\bibitem{wei2016zest}
H.~Wei, Y.~Wang, T.~Wo, Y.~Liu, and J.~Xu, ``Zest: a hybrid model on predicting
  passenger demand for chauffeured car service,'' in \emph{Proceedings of the
  25th ACM International on Conference on Information and Knowledge
  Management}, 2016, pp. 2203--2208.

\bibitem{wang2016traffic}
J.~Wang, Q.~Gu, J.~Wu, G.~Liu, and Z.~Xiong, ``Traffic speed prediction and
  congestion source exploration: A deep learning method,'' in \emph{Proceedings
  of the 16th IEEE International Conference on Data Mining (ICDM)}, 2016, pp.
  499--508.

\bibitem{tong2017simpler}
Y.~Tong, Y.~Chen, Z.~Zhou, L.~Chen, J.~Wang, Q.~Yang, J.~Ye, and W.~Lv, ``The
  simpler the better: a unified approach to predicting original taxi demands
  based on large-scale online platforms,'' in \emph{Proceedings of the 23rd ACM
  SIGKDD international conference on knowledge discovery and data mining},
  2017, pp. 1653--1662.

\bibitem{feng2018deepmove}
J.~Feng, Y.~Li, C.~Zhang, F.~Sun, F.~Meng, A.~Guo, and D.~Jin, ``Deepmove:
  Predicting human mobility with attentional recurrent networks,'' in
  \emph{Proceedings of the 2018 World Wide Web conference (WWW)}, 2018, pp.
  1459--1468.

\bibitem{yao2018deep}
H.~Yao, F.~Wu, J.~Ke, X.~Tang, Y.~Jia, S.~Lu, P.~Gong, J.~Ye, and Z.~Li, ``Deep
  multi-view spatial-temporal network for taxi demand prediction,'' in
  \emph{Proceedings of the AAAI Conference on Artificial Intelligence},
  vol.~32, no.~1, 2018.

\bibitem{wang2019origin}
Y.~Wang, H.~Yin, H.~Chen, T.~Wo, J.~Xu, and K.~Zheng, ``Origin-destination
  matrix prediction via graph convolution: a new perspective of passenger
  demand modeling,'' in \emph{Proceedings of the 25th ACM SIGKDD International
  Conference on Knowledge Discovery \& Data Mining}, 2019, pp. 1227--1235.

\bibitem{DBLP:conf/icde/WangTLXXL19}
Y.~Wang, Y.~Tong, C.~Long, P.~Xu, K.~Xu, and W.~Lv, ``Adaptive dynamic
  bipartite graph matching: {A} reinforcement learning approach,'' in
  \emph{Proceedings of the 35th {IEEE} International Conference on Data
  Engineering, {ICDE} 2019, Macao, China, April 8-11, 2019}, 2019, pp.
  1478--1489.

\bibitem{DBLP:conf/kdd/ShiTZSLY21}
D.~Shi, Y.~Tong, Z.~Zhou, B.~Song, W.~Lv, and Q.~Yang, ``Learning to assign:
  Towards fair task assignment in large-scale ride hailing,'' in \emph{The 27th
  {ACM} Conference on Knowledge Discovery and Data Mining, Virtual Event,
  Singapore, August 14-18, 2021}.\hskip 1em plus 0.5em minus 0.4em\relax {ACM},
  2021, pp. 3549--3557.

\bibitem{tong2019two}
Y.~Tong, Y.~Zeng, B.~Ding, L.~Wang, and L.~Chen, ``Two-sided online micro-task
  assignment in spatial crowdsourcing,'' \emph{IEEE Transactions on Knowledge
  and Data Engineering}, 2019.

\bibitem{tong2016mobile}
Y.~Tong, J.~She, B.~Ding, L.~Wang, and L.~Chen, ``Online mobile micro-task
  allocation in spatial crowdsourcing,'' in \emph{2016 IEEE 32Nd international
  conference on data engineering (ICDE)}.\hskip 1em plus 0.5em minus
  0.4em\relax IEEE, 2016, pp. 49--60.

\bibitem{tong2016online}
Y.~Tong, J.~She, B.~Ding, L.~Chen, T.~Wo, and K.~Xu, ``Online minimum matching
  in real-time spatial data: experiments and analysis,'' \emph{Proceedings of
  the VLDB Endowment}, vol.~9, no.~12, pp. 1053--1064, 2016.

\bibitem{qu2018product}
Y.~Qu, B.~Fang, W.~Zhang, R.~Tang, M.~Niu, H.~Guo, Y.~Yu, and X.~He,
  ``Product-based neural networks for user response prediction over multi-field
  categorical data,'' \emph{ACM Transactions on Information Systerm (TIS.},
  2018.

\bibitem{Rendle2010factorization}
S.~Rendle, ``Factorization machines,'' in \emph{2010 IEEE International
  Conference on Data Mining (ICDM)}, 2010, pp. 995--1000.

\bibitem{xiao2017attentional}
J.~Xiao, H.~Ye, X.~He, H.~Zhang, F.~Wu, and T.-S. Chua, ``Attentional
  factorization machines: Learning the weight of feature interactions via
  attention networks,'' in \emph{Proceedings of the 26th International Joint
  Conference on Artificial Intelligence (IJCAI)}, 2017.

\bibitem{tang2018ranking}
J.~Tang and K.~Wang, ``Ranking distillation: Learning compact ranking models
  with high performance for recommender system,'' in \emph{Proceedings of the
  24th ACM SIGKDD International Conference on Knowledge Discovery \& Data
  Mining}, 2018, pp. 2289--2298.

\bibitem{wang2020next}
Q.~Wang, H.~Yin, T.~Chen, Z.~Huang, H.~Wang, Y.~Zhao, and N.~Q. Viet~Hung,
  ``Next point-of-interest recommendation on resource-constrained mobile
  devices,'' in \emph{Proceedings of The Web Conference 2020}, 2020, pp.
  906--916.

\end{thebibliography}
\vspace{10pt}
\newpage
\begin{IEEEbiography}[{\includegraphics[width=1in,height=1.25in,clip,keepaspectratio]{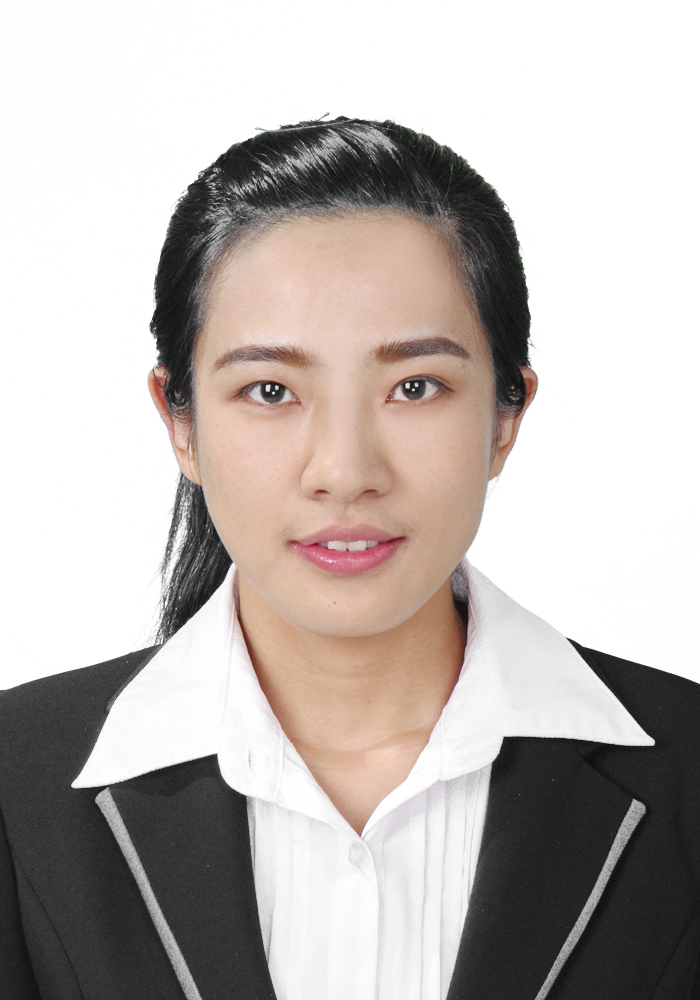}}]{Yuandong Wang} 
	is currently a PhD candidate with School of  Computer Science and Engineering at Beihang University. Her research interests include representation learning, traffic data mining and urban computing.
\end{IEEEbiography}

\vspace{-15pt}
\begin{IEEEbiography}[{\includegraphics[width=1in,height=1.25in,clip,keepaspectratio]{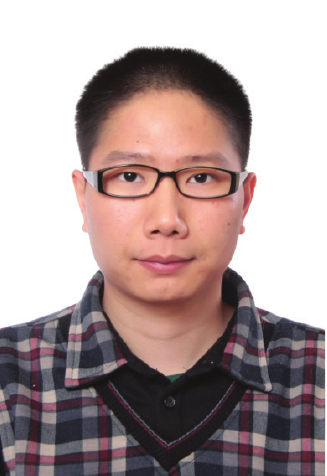}}]{Hongzhi Yin}
	received the Ph.D. degree in computer science from Peking University in 2014. He is an Associate Professor and Future Fellow with the University of Queensland. He received the Australia Research Council Future Fellowship
	and Discovery Early-Career Researcher Award in 2016 and 2021. His research interests include recommendation system, user profiling, topic models, deep learning, social media mining, and location-based services.
\end{IEEEbiography}

\vspace{-15pt}
\begin{IEEEbiography}[{\includegraphics[width=1in,height=1.25in,clip,keepaspectratio]{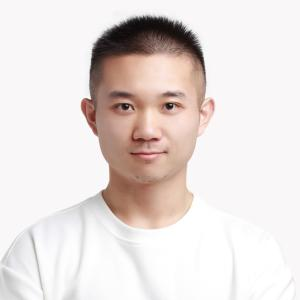}}]{Lian Wu} is currently a senior algorithm engineer in DiDi, working on order dispatching at the Transaction Engine Department. His research interests focus on the real-world applications of data mining and operation optimization through machine learning methods.
\end{IEEEbiography}

\vspace{-15pt}
\begin{IEEEbiography}[{\includegraphics[width=1in,height=1.25in,clip,keepaspectratio]{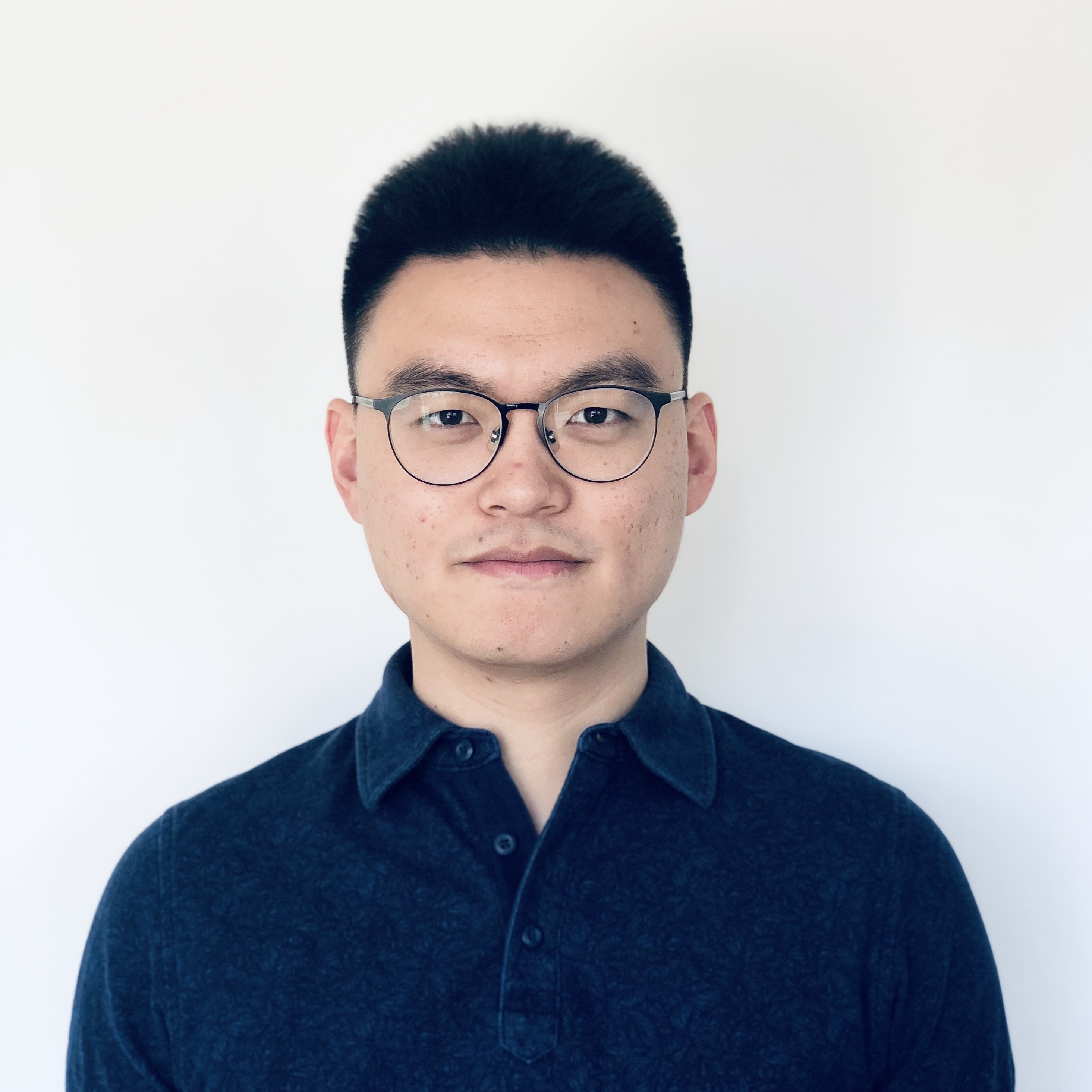}}]{Tong Chen}
	received his PhD degree in computer science from The University of Queensland in 2020. He is currently a postdoctoral research fellow with the Data Science research group, School of Information Technology and Electrical Engineering, The University of Queensland. His research interests include data mining, recommender systems, user behavior modeling and predictive analytics. 
\end{IEEEbiography}

\vspace{-15pt}
\begin{IEEEbiography}[{\includegraphics[width=1in,height=1.25in,clip,keepaspectratio]{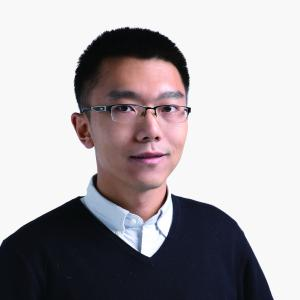}}]{Chunyang Liu}
	is a senior algorithm engineer in DiDi Chuxing. He received his Ph.D. from University of Technology, Sydney and B.Sc. from Shanghai Jiao Tong University. His main research interests include machine learning, urban computing and operation research.
\end{IEEEbiography}


\end{document}